\documentclass{article}

    \PassOptionsToPackage{numbers, compress}{natbib}

    \usepackage[final]{neurips_2022}

\usepackage[utf8]{inputenc} 
\usepackage[T1]{fontenc}    
\usepackage[hidelinks]{hyperref}       
\usepackage{url}            
\usepackage{booktabs}       
\usepackage{amsfonts}       
\usepackage{nicefrac}       
\usepackage{microtype}      
\usepackage{xcolor}         

\author{%
  Parnian Kassraie\\
  ETH Zurich\\
  \texttt{pkassraie@ethz.ch} \\
   \And
   Andreas Krause \\
ETH Zurich\\
   \texttt{krausea@ethz.ch} \\
   \And
   Ilija Bogunovic \\
   University College London \\
   \texttt{i.bogunovic@ucl.ac.uk} \\
}

\usepackage{answers}
\usepackage{setspace}
\usepackage{graphicx}
\usepackage{enumitem}
\usepackage{multicol}
\usepackage{mathrsfs}
\usepackage{amsmath,amsthm,amssymb,amsfonts, bm}

\usepackage{xspace, mathtools}
\usepackage{dsfont}
\usepackage{bbm}

\usepackage[ruled]{algorithm2e}
\SetKwInput{KwInput}{Input}   
\SetKwInput{KwOutput}{Output} 
\SetKwProg{init}{Initialize}{}{}

\usepackage[capitalize]{cleveref}

\DeclareMathOperator*{\argmax}{arg\,max}

\DeclareMathOperator*{\tr}{tr}
\DeclarePairedDelimiter\abs{\lvert}{\rvert}%
\DeclarePairedDelimiter\norm{\lVert}{\rVert}%
\makeatletter
\let\oldabs\abs
\def\abs{\@ifstar{\oldabs}{\oldabs*}}
\let\oldnorm\norm
\def\norm{\@ifstar{\oldnorm}{\oldnorm*}}
\makeatother

\makeatletter
\newcommand{\pushright}[1]{\ifmeasuring@#1\else\omit\hfill$\displaystyle#1$\fi\ignorespaces}
\newcommand{\pushleft}[1]{\ifmeasuring@#1\else\omit$\displaystyle#1$\hfill\fi\ignorespaces}
\makeatother

\newcommand{\R}{\mathbb{R}}

\def\vtheta{{\bm{\theta}}}
\def\vphi{{\bm{\phi}}}

\def\vb{{\bm{b}}}

\def\vf{{\bm{f}}}
\def\vg{{\bm{g}}}
\def\vh{{\bm{h}}}

\def\vk{{\bm{k}}}

\def\vw{{\bm{w}}}
\def\vx{{\bm{x}}}
\def\vy{{\bm{y}}}

\def\mA{{\bm{A}}}

\def\mG{{\bm{G}}}

\def\mI{{\bm{I}}}

\def\mK{{\bm{K}}}

\def\mP{{\bm{P}}}
\def\mQ{{\bm{Q}}}

\def\mW{{\bm{W}}}

\def\mZ{{\bm{Z}}}

\def\mPhi{{\bm{\Phi}}}

\def\sP{{\mathbb{P}}}

\def\sR{{\mathbb{R}}}
\def\sS{{\mathbb{S}}}

\def\cC{{\mathcal{C}}}

\def\cG{{\mathcal{G}}}
\def\cH{{\mathcal{H}}}

\def\cL{{\mathcal{L}}}

\def\cN{{\mathcal{N}}}
\def\cO{{\mathcal{O}}}

\newtheorem{theorem}{Theorem}[section]

\newtheorem{proposition}[theorem]{Proposition}
\newtheorem{lemma}[theorem]{Lemma}

\newcommand{\poly}{{\rm poly}}

\DeclareMathOperator{\polylog}{polylog}

\def\gnnucb{{\textsc{GNN-UCB}}\xspace}
\def\nnucb{{\textsc{NN-UCB}}\xspace}

\def\phasedalg{{\textsc{GNN Phased Elimination (\gnnus)}}\xspace}
\def\gnnus{{\textsc{GNN-PE}}\xspace}
\def\nnus{{\textsc{NN-PE}}\xspace}

\newcommand{\nn}{f_{\mathrm{NN}}}
\newcommand{\ntk}{{k_{\mathrm{NN}}}}
\newcommand{\ntklong}{{\kappa_{\mathrm{NN}}}}

\newcommand{\gnn}{f_{\mathrm{GNN}}}
\newcommand{\vgnn}{\vf_\mathrm{GNN}}
\newcommand{\ggnn}{\vg_{\mathrm{GNN}}}
\newcommand{\gradnn}{\vg_{\mathrm{NN}}}
\newcommand{\gntk}{{k_{\mathrm{GNN}}}}
\newcommand{\tgntk}{{\hat k_{\mathrm{GNN}}}}

\newcommand{\mgnn}{\bar{\mG}}
\newcommand{\mgntk}{\mK_{\mathrm{GNN}}}
\newcommand{\mgntkT}{\mK_{\mathrm{GNN},T}}

\def\vkhat{{\hat \vk}}
\def\mKhat{{\hat\mK}}

\newcommand{\Hgntk}{\cH_{\mathrm{GNN}}}

\usepackage{subfig}
\usepackage{colortbl}

\definecolor{TODO}{rgb}{0.88, 0.66, 0.37}
\definecolor{comment}{HTML}{4A7856}
\definecolor{blue_table}{HTML}{ECF5FE}

\title{Graph Neural Network Bandits}

\begin{document}

\maketitle

\begin{abstract}

    We consider the bandit optimization problem with the reward function defined over graph-structured data.
    This problem has important applications in molecule design and drug discovery, where the reward is naturally invariant to graph permutations. 
    The key challenges in this setting are scaling to large domains, and to graphs with many nodes.
    We resolve these challenges by embedding the permutation invariance into our model.
    In particular, we show that graph neural networks (GNNs) can be used to estimate the reward function, assuming it resides in the Reproducing Kernel Hilbert Space of a permutation-invariant additive kernel.
    By establishing a novel connection between such kernels and the graph neural tangent kernel (GNTK), we introduce the first GNN confidence bound and use it to design a phased-elimination algorithm with sublinear regret.
    Our regret bound depends on the GNTK's maximum information gain, which we also provide a bound for.
    While the reward function depends on all $N$ node features, our guarantees are {\em independent} of the number of graph nodes $N$.
    Empirically, our approach exhibits competitive performance and scales well on graph-structured domains. 
    \looseness -1

\end{abstract}
\vspace{0.5ex}
\section{Introduction}
\vspace{0.5ex}
\label{sec:intro}
Contemporary bandit optimization approaches consider problems on large or continuous domains and have successfully been applied to a significant number of machine learning and real-world applications, e.g., in mobile health, environmental monitoring, economics, and hyperparameter tuning, to name a few. The main idea behind them is to exploit the correlations between the rewards of “similar” actions. This in turn, has resulted in increasingly rich models of reward functions (e.g., in linear and kernelized bandits \cite{srinivas2009gaussian, chowdhury2017kernelized}), including several recent attempts to harness deep neural networks for bandit tasks (see, e.g, \cite{zhou2020neural, krause2011contextual}). A vast majority of previous works only focus on standard input domains and obtaining theoretical regret bounds. 

Learning on graph-structured data, such as molecules or biological graph representations, requires designing sequential methods that can effectively exploit the structure of graphs. Consequently, graph neural networks (GNNs) have received attention as a rapidly expanding class of machine learning models. They deem remarkably well-suited for prediction tasks in applications such as designing novel materials \cite{guo2020semi},  drug discovery \cite{jiang2021could}, structure-based protein function prediction \cite{gligorijevic2021structure}, etc. This rises the question of how to bridge the gap, and design bandit optimization algorithms on graph-structured data that can exploit the power of graph neural networks to approximate a graph reward.\looseness=-1


In this paper, we consider bandit optimization over graphs and propose to employ graph neural networks with one convolutional layer to estimate the unknown reward. To scale to \emph{large} graph domains (both in the number of graphs and number of nodes), we propose practical structural assumptions to model the reward function. In particular, we propose to use \emph{permutation invariant additive} kernels. We show a novel connection between such kernels and the \emph{graph neural tangent kernel} (GNTK) that we define in \cref{sec:gnn}. Our main result are \emph{GNN confidence bounds} that can be readily used in sequential decision-making algorithms to achieve sublinear regret bounds (see \cref{sec:regret}).\looseness=-1

\textbf{Related Work.} 
Our work extends the rich toolbox of methods for kernelized bandits and Bayesian optimization (BO) that work under the norm bounded Reproducing Kernel Hilbert Space assumption \cite{srinivas2009gaussian, de2012exponential, valko2013finite, chowdhury2017kernelized}). 
The majority of these methods are designed for general Euclidean domains and rely on kernelized confidence sets to select which action to query next. 
The exception is \cite{valko2014spectral}, that consider the spectral setting in which the reward function is a linear combination of the eigenvectors of the graph Laplacian and the bandit problem is defined over nodes of a single graph. 
In contrast, our focus is on optimizing over graph domains (i.e., set of graphs), and constructing confidence sets that can quantify the uncertainty of graph neural networks estimates.

This work contributes to the literature on \emph{neural bandits}, in which a fully-connected \citep{zhou2020neural,zhang2020neural,gu2021batched}, or a single hidden layer convolutional network \citep{kassraie2021neural} is used to estimate the reward function. These works provide sublinear cumulative regret bounds in their respective settings, however, when applied directly to graph features (as we demonstrate in \cref{sec:regret}), these approaches do not scale well with the number of graph nodes. 

Due to its important applications in molecule design, sequential optimization on graphs has recently received considerable attention. 
For example, in \cite{korovina2020chembo}, the authors propose a kernel to capture similarities between graphs, and at every step, select the next graph through a kernelized random walk. Other works (e.g., \citep{gomez2018automatic, griffiths2020constrained, jin2018junction,stanton2022accelerating}) encode graph representations to the (continuous) latent space of a variational autoencoder and perform BO in the latent space. While practically relevant for discovering novel molecules with optimized properties, these approaches lack theoretical guarantees and deem computationally demanding.\looseness=-1

A primary focus in our work is on embedding the natural structure of the data, i.e., permutation invariance, into the reward model. 
This is inspired by the works of \citep{bietti2021sample, mei2021learning} that consider invariances in kernel-based supervised learning.
Consequently, the graph neural tangent kernel plays an integral role in our theoretical analysis. \citet{du2019graph} provide a recursive expression
for the tangent kernel of a GNN, without showing that the obtained expression is the limiting tangent kernel as defined in \citet{jacot2018neural} (i.e., as in \cref{{eq:gntk_def}}). In contrast, we analyze the learning dynamics of the GNN and properties of the GNTK by exploiting the connection between the structure of a graph neural network and that of a neural network (in \cref{eq:gnn_vs_nn}). We recover that the graph neural tangent kernel also encodes additivity. Additive models for bandit optimization have been previously studied in \cite{kandasamy2015high} and \cite{rolland2018high}, however, these works only focus on Euclidean domains and standard base kernels. 
Finally, we build upon the recent literature on elimination-based algorithms that make use of maximum variance reduction sampling \citep{contal2013parallel, bogunovic2016truncated,bogunovic2021misspecified,bogunovic2022robust,vakili2021optimal,li2022gaussian}.
One of our proposed algorithms, \gnnus, employs a phased elimination strategy together with our GNN confidence sets.   

\textbf{Main Contributions.} \looseness -1 We introduce a bandit problem over graphs and propose to capture prior knowledge by modeling the unknown reward function using a permutation invariant additive kernel.
We establish a key connection between such kernel assumptions and the graph neural tangent kernel (\cref{lem:main_gntk_perminvar}).
By exploiting this connection, we provide novel statistical confidence bounds for the graph neural network estimator (\cref{thm:CI_gnn}). 
We further prove that a phased elimination algorithm that uses our GNN-confidence bounds (\gnnus) achieves sublinear regret (\cref{thm:gnnus_regret}). 
Importantly, our regret bound scales favorably with the number of graphs and is {\em independent} of the number of graph nodes (see \cref{tbl:summary}). 
Finally, we empirically demonstrate that our algorithm consistently outperforms baselines  across a range of problem instances.\looseness=-1


\vspace{0.5ex}
\section{Problem Statement}\label{sec:problem_statement}
\vspace{0.5ex}
We consider a bandit problem where the learner aims to optimize an \emph{unknown} reward function via sequential interactions with a stochastic environment.
At every time step $t \in \lbrace 1, \dots, T \rbrace$, the learner selects a graph $G_t$ from a graph domain $\cG$
and observes a noisy reward $y_t = f^*(G_t) + \epsilon_t$, where $f^*:\cG\rightarrow\sR$ is the reward function and $\epsilon_t$ is i.i.d.~zero-mean sub-Gaussian noise with known variance proxy $\sigma^2$. 
Over a time horizon $T$, the learner seeks a small \emph{cumulative} regret
$
    R_T = \sum_{t=1}^T f^*(G^*) - f^*(G_t),
$ where $G^* \in \argmax_{G \in \cG} f^*(G)$. 
The aim is to attain regret that is {\em sublinear} in $T$, meaning that $R_T/T\rightarrow 0$ as $T \to \infty$, which implies convergence to the optimal graph.
As an example application, 
consider drug or material design, where molecules may be represented with graph structures (e.g., from \textsc{Smiles} representations \citep{anderson1987smiles}) and the reward $f^*(G)$ can correspond to an unknown molecular property of interest, e.g., atomization energy.
Evaluating such properties typically requires running costly simulations or experiments with noisy outcomes. To identify the most promising candidate, e.g., the molecule with the highest atomization energy, molecules are sequentially recommended for testing and the goal is to find the optimal molecule with the least number of evaluations. \looseness=-1

\textbf{Graph Domain.} 
\looseness -1 We assume that the domain $\cG$ is a finite set of undirected graphs with $N$ nodes.\footnote{This assumption is for ease of exposition. 
Graphs with fewer than $N$ nodes can be treated by adding auxiliary nodes with no features that are disconnected from the rest of the graph.
}
Without exploiting structure, standard bandit algorithms (e.g., \cite{auer2002finite}) cannot generalize across graphs, and their regret linearly depends on $|\cG|$. 
To capture the structure, we consider reward functions depending on features associated with the graph nodes.
Similar to \citet{du2019graph}, we associate each node $j \in [N]$ with a feature vector $\vh_{G,j} \in \sR^{d}$, for every graph $G \in \cG$.
We use $\vh_{G} = (\vh_{G,j})_{j=1}^N\in \mathbb{R}^{Nd}$ to denote the concatenated vector
of all node features, and $\cN(j)$ as the neighborhood of node $j$, including itself.
We define the aggregated node feature $\bar \vh_{G,j} = \sum_{i \in \cN(j)} \vh_{G,i}/\vert\vert\sum_{i \in \cN(j)} \vh_{G,i}\vert\vert_2 $ as the normalized sum of the neighboring nodes' features.
Similarly, $\bar \vh_{G} \in \mathbb{R}^{Nd}$ denotes the aggregated features, stacked across all nodes. 
Lastly, we let $P_N$ be the group of all permutations of length $N$, and use $c \cdot G$ to denote a permuted graph, where a permutation $c \in P_N$ is a bijective mapping from $\{1, \dots, N\}$ onto itself. 
Permuting the nodes of a graph $c \cdot G$ produces a permuted feature vector $\vh_{c \cdot G} := (\vh_{G, c(j)})_{j=1}^N$, and the same holds for the aggregated features $\bar \vh_{c \cdot G}$ .\looseness=-1

\textbf{Reward Model.} Practical graph optimization problems, such as drug discovery and materials optimization often do not depend on how the graphs' nodes in the dataset are ordered.
We incorporate this structural prior into modeling the reward function, and consider functions that are \emph{invariant to node permutations}. We assume that $f^*$ depends on the graph only through the aggregated node features and gives the same reward for all permutations of a graph, i.e.,
$f^*(c\cdot G) = f^*(G)$, for any $G \in \cG$ and $c \in P_N$. 
To guarantee such an invariance, we assume that the reward belongs to the reproducing kernel Hilbert space (RKHS) corresponding to a permutation invariant kernel
\begin{equation*} \label{eq:perm_invariant_kbar}
    \bar k(G, G') =  \frac{1}{\vert P_N \vert^2} \sum_{c,c' \in P_N} k(\bar\vh_{c \cdot G},  \bar\vh_{c'\cdot G'}),
\end{equation*}
where $k$ can be any kernel defined on graph representations $\bar{\vh}_{G}$.
This assumption further restricts the hypothesis space to permutation invariant functions defined on $Nd$--dimensional vector representations of graphs. 
This is due to the reproducing property of the RKHS which allows us to write
$
f(G) = \langle f, \bar k(G, \cdot) \rangle = \langle f, \bar k(c \cdot G, \cdot) \rangle = f (c\cdot G)
$.
To make progress when optimizing over graphs with a \emph{large} number of nodes $N$, we assume that $k$ 
decomposes additively over node features, i.e., 
\begin{equation*}\label{eq:additive_k}
    k(\bar \vh_G, \bar \vh_{G'}) = \frac{1}{N} \sum_{j = 1}^N k^{(j)}(\bar\vh_{G,j}, \bar \vh_{G',j}).
\end{equation*}
Thereby, we obtain an \emph{additive} graph kernel that is \emph{invariant} to node permutations:
\begin{equation}\label{eq:perm_invar_def}
\bar k(G,G') = \frac{1}{\vert P_N \vert^2} \sum_{c,c' \in P_N} \frac{1}{N} \sum_{j = 1}^N k^{(j)}(\bar\vh_{G,c(j)}, \bar \vh_{G',c'(j)}).
\end{equation}
For an arbitrary choice of $k^{(j)}$, calculating $\bar k$ requires a costly sum over $(N!)^2$ operands, since $|P_N| = N!$. In \cref{sec:gnn}, we select a base kernel for which the sum can be reduced to $N^2$ terms.
We are now in a position to state our main assumption. We assume that $f^*$ belongs to the RKHS of $\bar k$ and has a $B$-bounded RKHS norm. The norm-bounded RKHS regularity assumption is typical in the kernelized and neural bandits literature \citep{srinivas2009gaussian, chowdhury2017kernelized, zhou2020neural, kassraie2021neural}.
Note that \cref{eq:perm_invar_def} only puts a structural prior on the kernel function, i.e., it describes the generic form of an additive permutation invariant graph kernel. Specifying the base kernels $k^{(j)}$ determines the representation power of $\bar k$. The smoother the base kernels are, the less complex the RKHS of $\bar k$ will be. In \cref{sec:gnn}, we set the base kernels $k^{(j)}$ such that $\bar k$ becomes the expressive graph neural tangent kernel.

\vspace{0.5ex}
\section{Graph Neural Networks} \label{sec:gnn}
\vspace{0.5ex}
Graph neural networks are effective models for learning complex functions defined on graphs.
As in \citet{du2019graph}, we consider graph networks that have a {\em single} graph convolutional layer and $L$ fully-connected ReLU layers of equal width $m$. Such a network $\gnn(G; \vtheta): \cG \rightarrow \R$ may be recursively defined as follows:\looseness=-1
\begin{equation}
    \label{eq:def_gnn}
    \begin{split}
        f^{(1)}(\bar\vh_{G, j}) &= \mW ^{(1)}\bar\vh_{G, j},\\
     f^{(l)}(\bar\vh_{G, j}) &= \sqrt{\frac{2}{m}} \mW^{(l)} \sigma_{\text{relu}}\big(f^{(l-1)}(\bar\vh_{G, j})\big) \in \R^m,1<l\leq L\\
     \gnn(G; \vtheta) &=  \frac{1}{N}\sum_{j=1}^N \sqrt{2} \mW^{(L+1)} \sigma_{\text{relu}}\big(f^{(L)}(\bar \vh_{G, j})\big),
    \end{split}
\end{equation}
where $\vtheta \coloneqq (\mW^{(i)})_{i\leq L+1}$ is initialized randomly with standard normal i.i.d.~entries, and $\sigma_{\text{relu}}(\vx) \coloneqq \max(\bm{0}, \vx)$. The network operates on aggregated node features $\bar\vh_{G, j}$ as typical in Graph Convolutional Networks \citep{kipf2016semi}.
For convenience, we assume that at initialization $\gnn(G; \vtheta^0) = 0$,  for all $G \in \cG$, similar to \citep{kassraie2021neural, zhou2020neural}. This assumption can be fulfilled without loss of generality, with a similar treatment as in \citep[Appendix B.2]{kassraie2021neural}.

\textbf{Embedded Invariances.} In this work, we use graph neural networks to estimate the unknown reward function $f^*$.
This choice is motivated by the expressiveness of the GNN, the fact that it scales well with graph size, and particularly due to the invariances embedded in its structure. 
We observe that the graph neural network $\gnn$ is invariant to node permutations, i.e., for all $G \in \cG$ and $c \in P_N$,
\[\gnn(G; \vtheta) = \gnn(c\cdot G; \vtheta).\]
The key step to show this property is proving that $\gnn$ can be expressed as an additive model of $L$-layer fully-connected ReLU networks,
\begin{equation*} \label{eq:gnn_vs_nn}
    \gnn(G; \vtheta) =  \frac{1}{N}\sum_{j=1}^N \nn(\bar{\vh}_{G,j}; \vtheta),
\end{equation*}
where $\nn$ has a similar recursive definition as $\gnn$ (see Equation~\ref{eq:def_nn}).
The above properties are formalized in \cref{lem:main_nn_gnn} and \cref{prop:gnn_invariance}.

\textbf{Lazy (NTK) Regime.} We initialize and train $\gnn$ in the well-known lazy regime \citep{chizat2019lazy}.
In this initialization regime, when the width $m$ is large, training with gradient descent using a small learning rate causes little change in the network's parameters.
Let $\ggnn(G, \vtheta) = \nabla_\vtheta\gnn(G,\vtheta)$ denote the gradient of the network. It can be shown that during training, for all $G \in \cG$, the network remains close to $\gnn(G, \vtheta^0) + \ggnn^T(G, \vtheta^0)(\vtheta - \vtheta^0)$, that is, its first order approximation around initialization parameters $\vtheta^0$. 
Training this linearized model with a squared error loss is equivalent to kernel regression with a {\em tangent kernel}
$
   \tilde k_{\mathrm{GNN}} (G, G')  := \ggnn^T(G; \vtheta^0)\;\ggnn(G'; \vtheta^0)\label{eq:finite_gntk}
$.
For networks of finite width, this kernel function is random since it depends on the random network parameters at initialization. We show in \cref{lem:main_gntk_ntk}, that in the infinite width limit, the tangent kernel converges to a deterministic kernel. This proposition introduces the Graph Neural Tangent Kernel as the limiting kernel, and links it to the Neural Tangent Kernel (\cite{jacot2018neural}, also defined in \cref{app:ntk_gntk}). \looseness-1
\begin{proposition}\label{lem:main_gntk_ntk}
Consider any two graphs $G$ and $G'$ with $N$ nodes and $d$-dimensional node features. In the infinite width limit, the tangent kernel $\tilde k_{\mathrm{GNN}} (G, G')/m$ converges to a deterministic kernel,
\begin{equation}\label{eq:gntk_def}
\gntk(G, G') \coloneqq \lim_{m \rightarrow \infty} \tilde k_{\mathrm{GNN}} (G, G')/m.
\end{equation}
which we refer to as the Graph Neural Tangent Kernel (GNTK). Moreover, 
\begin{equation}\label{eq:gntk_vs_ntk}
        \gntk(G, G')= 
        \frac{1}{N^2}\sum_{j,j'=1}^N \ntk(\bar\vh_{G, j}, \bar \vh_{G',j'})
\end{equation}
where $\ntk: \sS^{d-1} \times \sS^{d-1}\rightarrow\sR$ is the Neural Tangent Kernel.
\end{proposition}
The proof is given in \cref{app:gnn_gntk_props}.
We note that $\bar\vh_{G,j}$ lies on the $d$-dimensional sphere, since the aggregated node features are normalized. The NTK is bounded by $1$ for any two points on the sphere \citep{bietti2020deep}. Therefore, \cref{lem:main_gntk_ntk} implies that the GNTK is also bounded, i.e., $\gntk(G, G')\leq 1$ for any $G, G' \in \cG$.
This proposition yields a kernel which captures the behaviour of the lazy GNN.
While defined on graphs with $Nd$ dimensional representations, the effective input domain of this kernel is $d$-dimensional. This advantage directly stems from the additive construction of the GNTK.
The next proposition uncovers the embedded structure of the GNTK by showing a novel connection between the GNTK and $\bar k$, the permutation invariant additive kernel from \cref{eq:perm_invar_def}.
The proof is presented in \cref{app:gnn_gntk_props}. \looseness-1

\begin{proposition}\label{lem:main_gntk_perminvar} Consider $\bar k$ from \cref{eq:perm_invar_def}, where for every $1 \leq j \leq N$ the base kernel $k^{(j)}$ is set to be equal to $\ntk$,\looseness=-1
\[
\bar k(G,G') = \frac{1}{\vert P_N \vert^2} \sum_{c,c' \in P_N} \frac{1}{N} \sum_{j = 1}^N \ntk(\bar\vh_{G,c(j)}, \bar \vh_{G',c'(j)}).
\]
Then the permutation invariant additive kernel and the GNTK are identical, i.e., for all $G, \,G'\in \cG$,
\[
\bar k(G,G') = \gntk(G,G').
\]
\end{proposition}
This result implies that $\gntk$ inherits the favorable properties of the permutation invariant additive kernel class.
Hence, functions residing in $\Hgntk$, the RKHS of $\gntk$, are additive, invariant to node permutations, and act on $G$ through its aggregated node features.
While we use the GNTK as an analytical tool, this kernel can be of independent interest in kernel methods over graph domains.
In particular, calculating $\gntk$ requires significantly fewer operations compared to a kernel $\bar k$ with an arbitrary choice of $k^{(j)}$, for which calculating $\bar k(G,G')$  requires super-exponentially many operations in $N$ (See \cref{eq:perm_invar_def}). In contrast, due to the decomposition in \cref{eq:gntk_vs_ntk}, calculating $\gntk$ only costs a quadratic number of summations.\looseness=-1

\vspace{0.5ex}
\section{GNN Bandits} \label{sec:gnnucb}
\vspace{0.5ex}

The bandit literature is rich with algorithms that effectively balance exploration and exploitation to achieve sublinear regret. Two components are common in kernelized bandit optimization algorithms.
The {\em maximum information gain}, for characterizing the worst-case complexity of the learning problem \citep{srinivas2009gaussian,kandasamy2015high,chowdhury2017kernelized,vakili2021optimal};
and {\em confidence sets}, for quantifying the learner's uncertainty \citep{auer2008near, thompson1933likelihood,russo2014learning,chowdhury2017kernelized,lu2019information}.
Our first main result is an upper bound for the maximum information gain when the hypothesis space is $\Hgntk$ (\cref{thm:MIG_gntk}). We then propose valid confidence sets that utilize GNNs in \cref{{thm:CI_gnn}}.
These theorems may be of independent interest, as they can be used towards bounding the regret for a variety of bandit algorithms on graphs. Lastly, we introduce the \gnnus algorithm, together with its regret guarantee. \looseness-1
\vspace{0.5ex}
\subsection{Information Gain}
\vspace{0.5ex}
\looseness -1 In bandit tasks, the learner seeks actions that give a large reward while, at the same time, provide information about the unknown reward function.
The speed of learning about $f^*$ is commonly quantified via the maximum information gain.
Assume that the learner chooses a sequence of actions $(G_1, \dots, G_T)$ and observes noisy rewards, where the noise is i.i.d.~and drawn from a zero-mean sub-Gaussian distribution with a variance proxy $\lambda$.
The information gain of this sequence calculated via the GNTK is \looseness=-1
\[
I(G_1, \dots, G_T ; \gntk) =  \frac{1}{2}\log\det(\mI + \lambda^{-1} \mgntkT)
\]
with the kernel matrix $\mgntkT = [\gntk(G_i, G_j)]_{i,j \leq T}$.
The maximum information gain (MIG) \citep{srinivas2009gaussian} is then defined as: 
\begin{equation}\label{eq:MMI_def}
\gamma_{\mathrm{GNN},T} =  \max_{\substack{(G_1, \cdots, G_T)\\\forall t:G_t \in \cG}}  I(G_1, \cdots, G_T ; \gntk).
\end{equation}
In \cref{sec:regret}, we express regret bounds in terms of this quantity, as common in kernelized and neural bandits. 
In \cref{thm:MIG_gntk} we obtain a data-independent bound on the MIG. The proof is given in \cref{app:infogain}.\looseness=-1
\begin{theorem}[GNTK Information Gain Bound]\label{thm:MIG_gntk}
Suppose the observation noise is i.i.d., and drawn from a zero-mean sub-Gaussian distribution, and the input domain is $\cG$. Then the maximum information gain associated with $\gntk$ is bounded by
\[
\gamma_{\mathrm{GNN}, T} = \cO\left(T^{\frac{d-1}{d}}\log^{\frac{1}{d}}T\right).
\]
\end{theorem}
 We observe that the obtained MIG bound {\em does not} depend on $N$ the number of nodes in the graphs.
To highlight this advantage, we compare \cref{thm:MIG_gntk} to the equivalent bound for the vanilla neural tangent kernel which ignores the graph structure.
We consider the neural tangent kernel that operates on graphs through the $Nd$-dimensional vector of aggregated node features $\bar \vh_G$, \looseness=-1
\begin{equation}\label{eq:def_longntk}
 \ntklong(G, G') = \ntklong \left(\frac{\bar \vh_G}{N}, \frac{\bar\vh_{G'}}{N}\right).
\end{equation}
For $\ntklong$ the maximum information gain scales as
$
\gamma_{\mathrm{NN},T} = \cO(T^{(Nd-1)/Nd}\log^{1/Nd}T)
$, where $N$ appears in the exponent \citep{kassraie2021neural}. This results in poor scalability with graph size in the bandit optimization task, as we further demonstrate in \cref{sec:regret}. \cref{tbl:summary} summarizes this comparison.
\vspace{0.5ex}
\subsection{Confidence Sets}
\vspace{0.5ex}
Quantifying the uncertainty over the reward helps the learner to guide exploration and balance it against exploitation.
Confidence sets are an integral tool for uncertainty quantification.
Conditioned on the history $H_{t-1} = (G_i, y_i)_{i < t}$, for any $G \in \cG$, the set $\cC_{t-1}(G,\delta)$ defines an interval to which $f^*(G)$ belongs with a high probability such that,
 \begin{equation}\label{eq:def_conf_set}
 \sP\left(\forall G\in\cG:f^*(G) \in \cC_{t-1}(G, \delta)\right)\geq 1-\delta.
 \end{equation}
An approach common to the kernelized bandit literature is to construct sets of the form 
$
    \cC_{t-1}(G, \delta) = [\mu_{t-1}(G)\pm\beta_t\sigma_{t-1}(G)]
$
where $\beta_t$ depends on the confidence level $\delta$. 
The center of the interval, characterized by $\mu_{t-1}(\cdot)$, is the estimate of the reward, and the width $\beta_t \sigma_{t-1}(\cdot)$, reflects the uncertainty.
In this work, we utilize GNNs for construction of such sets.
To this end, we train a graph neural network to estimate the reward. We use the gradient of this network at initialization to approximate the uncertainty over the reward, as in \citep{zhou2020neural}.
    Let $ \gnn(G; \vtheta^{(J)}_{t-1} ) $ be the GNN trained with gradient descent for $J$ steps and by using learning rate $\eta$ on the loss
\[
\cL(\vtheta) = \frac{1}{t}\sum_{i<t} \big(\gnn(G_i, \vtheta)- y_i\big)_2^2  + m \lambda \norm{\vtheta-\vtheta^0}_2^2,
\]
where $\lambda$ is the regularization coefficient, and $\vtheta^0$ the network parameters at initialization.
We propose confidence sets of the form
  \[\cC_{t-1}(G, \delta) = [\hat \mu_{t-1}(G)\pm\beta_t\hat\sigma_{t-1}(G)],\]
where the center and width of the set are calculated via,
\begin{equation}\label{eq:GNNposteriors}
\begin{split}
     \hat\mu_{t-1}(G) & := \gnn(G; \vtheta_{t-1}^{(J)}),\\
        \hat\sigma^2_{t-1}(G) & := \frac{\ggnn^T(G; \vtheta^0)}{\sqrt{m}}\Big(\lambda \mI + \frac{1}{t}\sum_{i = 1}^{t-1} \frac{\ggnn^T(G_i; \vtheta^0)\ggnn(G_i; \vtheta^0)}{m}\Big)^{-1}\frac{\ggnn(G; \vtheta^0)}{\sqrt{m}}.
\end{split}
\end{equation}
Here $\ggnn(G; \vtheta^0) = \nabla_\vtheta \gnn(G; \vtheta^0)$ denotes the gradient at initialization.
Moreover, we use $\lambda_0 := \lambda_{\min}(\mgntk) >0$ to denote the minimum eigenvalue of the kernel matrix calculated for the entire domain, i.e., $\mgntk = [\gntk(G, G')]_{G, G' \in \cG}$. 
\cref{thm:CI_gnn} shows that this construction gives valid confidence intervals, i.e., it satisfies \cref{eq:def_conf_set}, when the reward function lies in $\Hgntk$ and has a bounded RKHS norm. 
 \begin{theorem}[GNN Confidence Bound] \label{thm:CI_gnn}
Set $\delta \in (0,1)$.
Suppose $f^* \in \cH_{\gntk}$ with a bounded norm $\norm{f^*}_{\gntk}\leq B$.
Assume that the random sequences $(G_i)_{i < t}$ and $(\epsilon_i)_{i<t}$ are statistically independent. 
Let the width
$
         m =   \poly \left(t, L, B, \vert \cG\vert,\lambda, \lambda_0^{-1}, \log( N/\delta)\right),
$
 learning rate $\eta = C(Lm+m\lambda)^{-1}$ with some universal constant $C$, and $J \geq 1$. Then for all graphs $G \in \cG$, with probability of at least $1-\delta$, 
\begin{equation*}
        \vert  f^*(G) -\hat\mu_{t-1}(G)\vert \lessapprox \beta_t \hat\sigma_{t-1}(G),
\end{equation*}
where $\hat \mu_{t-1}$ and $\hat \sigma_{t-1}$ are defined in \cref{eq:GNNposteriors} and
\[
\beta_t \approx \sqrt{2}B + \frac{\sigma}{\sqrt{\lambda}}\sqrt{2\log2\vert \cG\vert/\delta} .
\]
 \end{theorem}
The "$\approx$" notation in \cref{thm:CI_gnn} omits the terms that vanish with $t$, i.e., are $o(1)$. An exact version of the theorem without the aforementioned approximations is given in \cref{app:main_CI_proof}.


\begin{table*}[!t]
    \begin{center}
        \begin{tabular}{c|c|c}
            \hline 
            \textbf{Setting}  & \textbf{MIG Bound, $\gamma_{T}$} & \textbf{Cumulative Regret (Phased Elimination)}    \tabularnewline
            \hline 
            \hline
            Neural & $\cO\left(T^{\frac{Nd-1}{Nd}}\log^{\frac{1}{Nd}}T\right)$ &  $\tilde\cO\left(T^{\frac{2Nd-1}{2Nd}}\log^{\frac{1}{2Nd}}T\right)$  \tabularnewline
            \hline 
            \rowcolor{blue_table}
            Graph Neural &$\cO\left(T^{\frac{d-1}{d}}\log^{\frac{1}{d}}T\right)$ &  $\tilde\cO\left(T^{\frac{2d-1}{2d}}\log^{\frac{1}{2d}}T\right)$ \tabularnewline
        \end{tabular}
        \vspace{1ex}
        \caption{Summary of main bounds for the NN and GNN approaches. Here $T$ denotes the bandit horizon, $N$ the number of nodes in each graph, and $d$ the dimension of node features. The GNN guarantees are {\em independent} of $N$. \label{tbl:summary}}
    \end{center}
\end{table*}
\subsection{Bandit Optimization with Graph Neural Networks} \label{sec:regret}
\vspace{0.5ex}
The developed confidence sets can be used to assist the learner with controlling the growth of regret.
In this section, we give a concrete example on how our GNN confidence sets (Equation~\ref{eq:GNNposteriors}) can be used by an algorithm to solve bandit optimization tasks on graphs. 

We introduce GNN-Phased Elimination (\gnnus; see \cref{alg:gnnalg}) that consists of episodes of pure exploration over a set of plausible maximizer graphs, similar to \citep{bogunovic2021misspecified, li2022gaussian}. 
Each episode is followed by an elimination step, that makes use of GNN confidence bounds to shrink the set of plausible maximizers.
More formally, at step $t$ during an episode $e$, the learner selects actions via $G_{e,t} = \argmax_{G \in \cG_e} \hat \sigma_{e,t-1}(G)$, where $\cG_e$ is the set of graphs that might maximize $f^*$ according to the learner's current knowledge.
Once the episode is over after $T_e$ steps, the set $\cG_e$ is updated to contain graphs that still have a chance of being a maximizer according to the confidence bounds $[\hat \mu_{e,T_e}(G) \pm \beta_{T_e}\hat\sigma_{e,T_e}(G)]$ where $\hat \mu_{e,T_e}$ and $\hat \sigma_{e,T_e}$ are only computed based on the points within episode $e$.\looseness=-1

\cref{thm:gnnus_regret} shows that \gnnus incurs a sublinear control over the cumulative regret.  We provide the proof in \cref{app:gnnus_regret}. We use $\tilde{\mathcal{O}}(\cdot)$ notation to hide $\polylog(T)$ factors.
\begin{theorem}\label{thm:gnnus_regret}
Set $\delta \in (0,1)$.
Suppose $f^* \in \cH_{\gnn}$ with a bounded norm $\norm{f^*}_{\gntk}\leq B$.
Let the width
$
         m =   \poly \left(t, L, B, \vert \cG\vert,\lambda, \lambda_0^{-1}, \log( N/\delta)\right),
$
 learning rate $\eta = C(Lm+m\lambda)^{-1}$ with some universal constant $C$, and $J \geq 1$. Then with probability at least $1-\delta$, 
 \gnnus satisfies
\begin{align*}
    R_T =\tilde\cO \left(\sqrt{T\gamma_{T, \mathrm{GNN}}}\left( B + \frac{\sigma}{\sqrt{\lambda}}\sqrt{ \log \vert \cG\vert/\delta}\right)  \right).
\end{align*}
\end{theorem}
We can observe the benefit of working with a graph neural network by comparing the bound in \cref{thm:gnnus_regret} with the regret for a structure-agnostic algorithm.
Recall $\ntklong$ the vanilla NTK, defined over the concatenated feature vectors (Equation~\ref{eq:def_longntk}). For the sake of this comparison, we ignore the geometric structure and assume that $f^* \in \cH_{\mathrm{NN}}$. Swapping out $\gntk$ for $\ntklong$, and respectively the GNN with an NN as defined in \cref{eq:def_nn}, we obtain \nnus, the neural network counterpart of \gnnus. This algorithm accepts $Nd$-dimensional input vectors as actions. Similar to \cref{thm:gnnus_regret}, we can show that \nnus can satisfy a guarantee of $\cO\big(T^{(2Nd-1)/2Nd}\log^{1/2Nd}T\big)$ for the regret. This bound suggests that as $N$ grows, finding the optimal graph can become more challenging for the learner.
Working with $\gntk$ to encode the structure of the bandit problem, and consequently using the GNN to solve it, removes the dependency on $N$ in the exponent. 
This result is summarized in \cref{tbl:summary}. \looseness -1

We provide some intuition on why working with a permutation invariant model is beneficial for bandit optimization on graphs. 
Confidence sets which are constructed for member of $\cH_{\mathrm{NN}}$ are larger, and result in sub-optimal action selection.
Further, training the neural network is a more challenging task, since permutation invariance is not hard coded in the network architecture and has to be learned from the data. This results in less accurate reward estimates.
We refer the reader to \cref{app:shrink} for a more rigorous discussion. There we compare $\Hgntk$ and $\cH_{\mathrm{NN}}$, the hypothesis spaces corresponding to the two models, through the Mercer decomposition of their kernels.

\section{Experiments}\label{sec:experiments}

\looseness-1 We create synthetic datasets which may be of independent interest and can be used for evaluating and benchmarking machine learning algorithms on graph domains. Each dataset is constructed from a finite graph domain together with a reward function.
The domains are generated randomly and differ in properties of the member graphs that influence the problem complexity, e.g., number of nodes and edge density. Each domain $\cG_{p, N}$ consists of Erd\H{o}s-R\'{e}nyi random graphs, where each graph has $N$ nodes, and between each two nodes there exists an edge with probability $p$. The node features are i.i.d. $d=10$ dimensional standard Gaussian vectors. We choose $N \in \{5, 20, 100\}$,  $p \in \{ 0.05, 0.2, 0.95\}$, and thereby sample a total of $9$ different domains each containing $10000$ graphs. For instance, $\cG_{0.05, 5}$ denotes the domain with sparse and small graphs, while $\cG_{0.95, 100}$ is the domain of dense graphs with many nodes.
For every domain, we sample a random reward function $f:\cG_{p,N }\rightarrow \sR$ that is invariant to node permutations. We use $\mathrm{GP}(0, \gntk)$ as a prior, and sample $f$ from its posterior GP. The posterior is calculated using a small random dataset $(G_i, y_i)_{i \leq 5}$, where $y_i$ are drawn independently from $\cN(0,1)$ and $G_i$ are randomly chosen from $\cG_{p,N}$. 
The corresponding dataset is then $D_{p,N} = \{\left(G_i, f(G_i)\right)\vert G_i \in \cG_{p,N}\}$.\looseness-1


\textbf{Experiment Setup.}
Every performance curve in the paper shows an average over $20$ runs of the corresponding bandit problem, each with a different action set sampled from $\cG_{p,N}$. The shaded areas in all figures show the standard error across runs. 
In all experiments, the reward is observed with a zero-mean Gaussian noise of variance $\sigma=10^{-2}$. 
We always set width $m=2048$ and layers $L=2$, for every type of network architecture.
Four algorithms appear in our experiments. In addition to our main algorithm \gnnus, we introduce \gnnucb, which selects actions via $G_t = \argmax \hat\mu_{t-1}(G)+ \beta_t \hat\sigma_{t-1}(G)$, the classic UCB policy based on the GNN confidence sets. The pseudo-code is given in \cref{app:ucb_algs}. \nnucb, introduced by \citep{zhou2020neural}, is the neural counterpart of \gnnucb, and \nnus as discussed in \cref{sec:regret}.
To configure these algorithms, we only tune $\lambda$ and $\beta = \beta_t$, and we do so by using the simplest dataset $D_{0.05,5}$.
We find that the algorithms are not sensitive to domain configurations and work for all $D_{p,N}$ out of the box. 
Therefore, the same values for $\lambda$ and $\beta$ are used across all experiments. 
We include the complete result of our hyperparameter search in Figure~\ref{fig:hyper_d10}.

\textbf{Lazy training.}
We initialize the graph neural networks (and the NNs) in the lazy regime as described in \cref{eq:def_gnn} (and Eq.~\ref{eq:def_nn}).
Training a network in this regime with gradient descent causes little change in the weights. Consequently, it is challenging to effectively train a lazy network in practice.
Therefore, the stopping criterion for gradient descent plays a crucial role in achieving sublinear regret. Inaccurate estimation of the reward function disturbs the balance of exploration and exploitation, and leads the learner to poor optima.
To prevent this issue, we devise a stopping criterion that depends on the history $H_{t-1}$, such that, as $t$ grows, the network is often trained for more gradient descent steps $J$. 
This criterion can be employed by any neural bandit algorithm and may be of independent practical interest. The details of training with gradient descent, stopping and batching are given in \cref{app:training}.\looseness=-1 

\begin{figure}
    \centering
    \includegraphics[width = 0.95\linewidth]{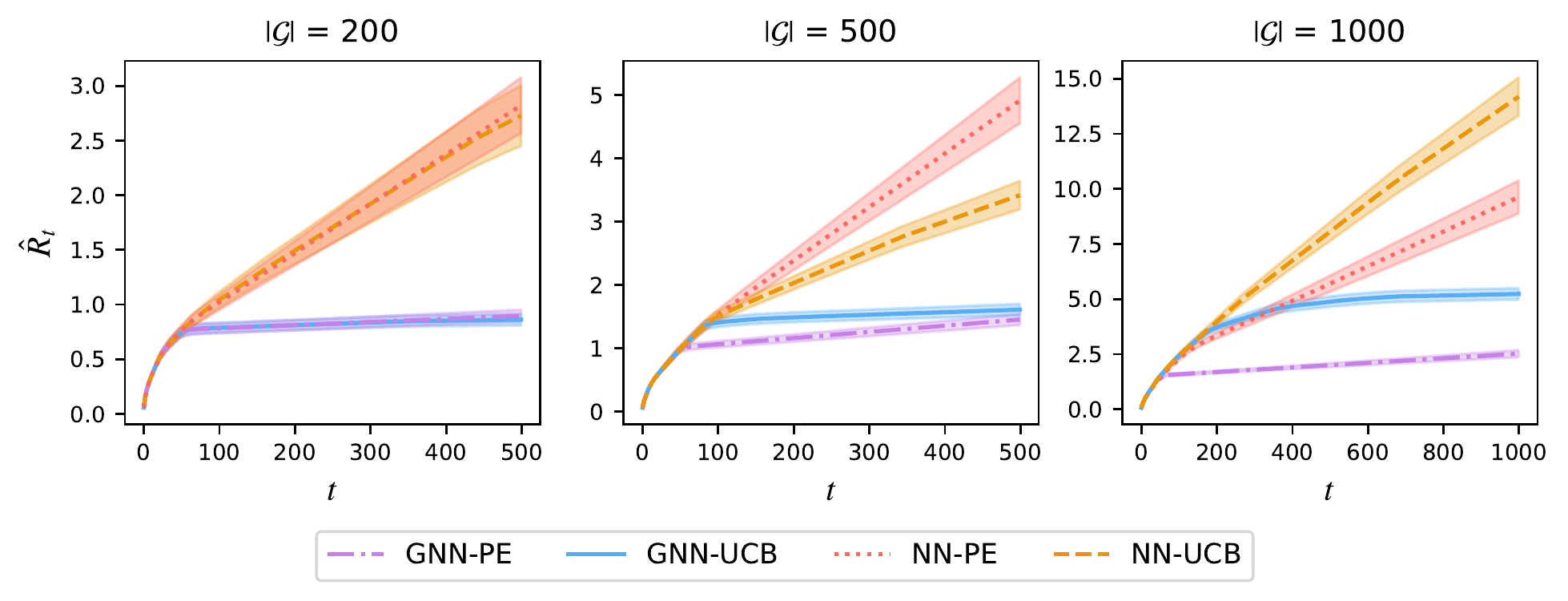}
    \caption{\label{fig:performance} Regret $\hat R_T$ over a time horizon of $500$ and $1000$ steps with $N=20$ and $p=0.2$. \gnnus consistently outperforms other algorithms and scales well with size of the action set $\vert \cG\vert$.}
\end{figure}

\textbf{Regret Experiments.}
\looseness -1 We assess the performance of the algorithms on bandit optimization tasks over different domains. In Figure~\ref{fig:performance}, we show the inference cumulative regret $\hat R_T = \sum_{t\leq T} f^*(G^*)- \max_{G\in \cG} \hat\mu_{t-1}(G)$,
for which we select graph domains with $N=20$ nodes and edge probability $p=0.2$. Figure~\ref{fig:benchmark} shows the regret for all dataset configurations. 
To verify scalability with $\vert \cG\vert$, we run the algorithms on action sets of increasing size $\vert \cG\vert \in \{200, 500, 1000\}$. Figure~\ref{fig:performance} presents the results: \gnnus consistently outperforms the other methods. It is evident that the algorithms built with GNN confidence sets find the optimal graph, regardless of the size of the domain. The GNN algorithms exhibit competitive performance, and attain sublinear regret for all dataset configurations.
The neural methods however, may fail to scale and find the optima in limited time.\looseness=-1

\begin{figure}[t!]
\begin{minipage}[b]{0.34\linewidth}
\centering
\includegraphics[width = \linewidth]{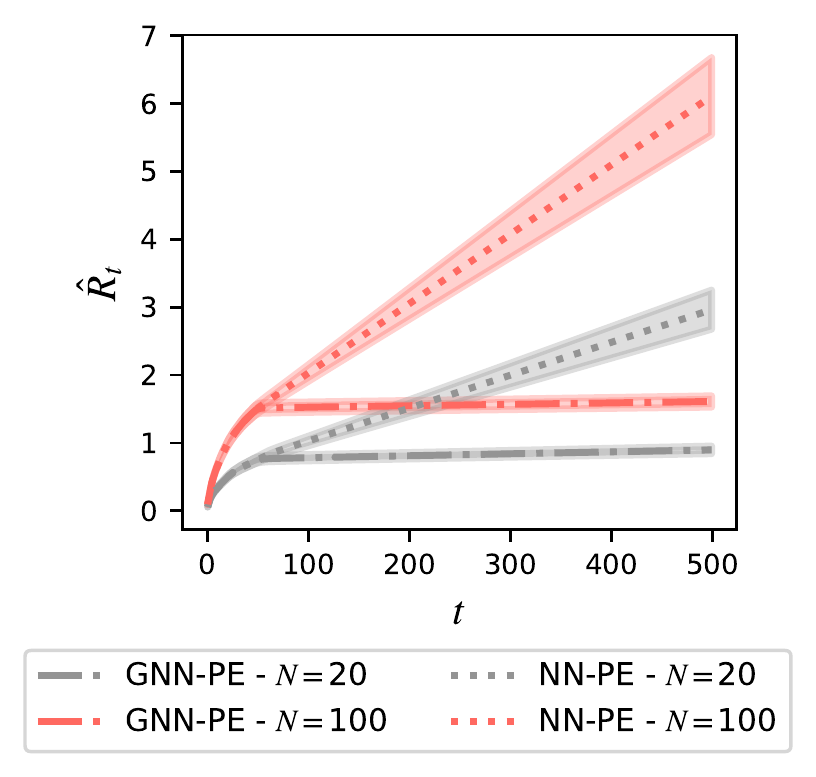}%
\caption{\label{fig:N_effect}\looseness-1 Increasing $N$ has little effect on \gnnus. }
\end{minipage}
\quad\quad\quad
\begin{minipage}[b]{0.6\linewidth}
\centering
\includegraphics[width = \linewidth]{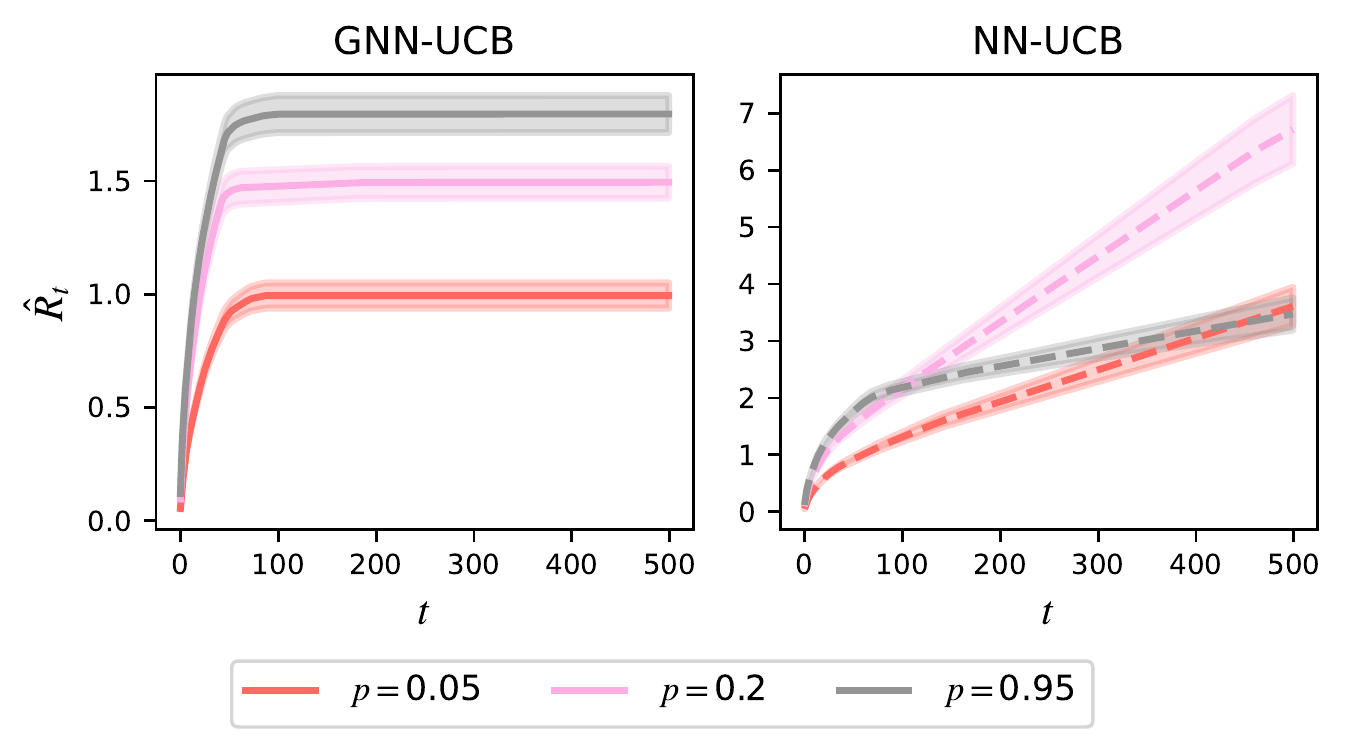}%
\caption{\label{fig:density_main} \looseness-1 Increasing the edge density of the graphs reduces the performance gap between \gnnucb and \nnucb.}
\end{minipage}
\end{figure}

\textbf{Scalability with Graph Size.}
In \cref{sec:regret}, we argue that using a neural network which takes $\bar \vh_G \in \sR^{Nd}$ as the input, causes the regret to grow with $\cO(T^{(2Nd-1)/2Nd})$.
The additive structure of the GNN, however, allows the learner to work on a $d$-dimensional domain, independent of graph size. Figure~\ref{fig:N_effect} reflects this behaviour. 
Fixing $p=0.2$, and $\vert \cG\vert =200$, we run the algorithm over domains with two graph sizes $N \in \{20, 100\}$. \gnnus achieves sublinear regret in both cases, and manages to find a global maxima within roughly the same number of steps. This is in contrast to \nnus, which is more affected by increasing graph size.
A similar comparison for all configurations and algorithms is plotted in Figure~\ref{fig:scalability}, and the same behaviour is observed across all settings: the performance of GNN methods scales well with $N$, while this is not the case for NN methods.\looseness-1


\textbf{Effect of Graph Density.}
As a final observation, we discuss the effect of edge density. 
Consider a complete graph $G$ with $\binom{N}{2}$ edges. 
The neighborhoods are symmetric and the aggregated node features $\bar\vh_{G,i}$ are identical for all $i \leq N$. 
Permutations on this graph will not change the output of either $\gnn$ or $\nn$.
Therefore, we expect that for dense graphs, i.e., large values of $p$, using a permutation invariant model comes with fewer benefits for the learner.
This is opposed to when the graph is sparse and the neighborhoods are asymmetric.
To verify this conjecture, we fix $N=20$, $\vert \cG \vert = 200$, and run the algorithm over domains with graphs of different edge probability $p \in \{ 0.05, 0.2, 0.95\}$. Figure~\ref{fig:density_main} shows that while \gnnucb always achieves sublinear regret, it takes longer to find the optima when the graphs are more dense.
\nnucb however, improves as the edge probability $p$ grows, since, roughly put, the graphs in the domain are becoming invariant to permutations. Therefore Figure~\ref{fig:density_main} confirms that the performance gap between the two method is reduced for graphs that are more dense.
In Figure~\ref{fig:density}, we plot the effect of graph density for other dataset configurations and bandit algorithms. This behaviour is observed predominantly for the UCB algorithms. \looseness -1


\section{Conclusion} \label{sec:conclusion}

We analyze the use of graph neural networks in bandit optimization tasks over large graph domains. The main takeaway is that encoding the natural structure of the environment into the model, reduces the complexity of the task for the learner. 
By selecting a kernel that embeds invariances, we introduce structure into the algorithm in a principled manner.
Importantly, we propose key structural assumptions on the graph reward function and establish a novel connection between additive permutation invariant kernels and the GNTK.
We construct valid graph neural network confidence sets, and use it to build a GNN bandit algorithm that achieves sublinear regret.
While all node features contribute to the graph's reward, our bounds are independent of the number of nodes.  
This result holds for GNNs with a single convolutional layer and graphs with node feature representation. An immediate next step is to generalize this approach to
other more complex graph neural network architectures and representations (e.g., by including information about graph edges) and investigating their effectiveness for bandit optimization.
Our analysis opens up two avenues of future research.
The proposed kernel and the graph confidence sets may be used in other algorithms for sequential decision-making tasks on graphs. Additionally, our approach of embedding the environment's permutation invariant structure into the algorithm may inspire further work on structured bandit optimization in presence of invariances. \looseness -1

\begin{ack}
We thank Jonas Rothfuss for his valuable suggestions regarding the experiments. 
We acknowledge Deepak Narayanan's effort on an earlier version of the code.
We thank Nicolas Emmenegger and Scott Sussex for their thorough feedback, and lastly, we thank Alex H{\"a}gele for fruitful discussions regarding the writing.  
This research was supported by the European Research Council (ERC) under the European Union’s Horizon 2020 research and Innovation Program Grant agreement no. 815943. 
\end{ack}

\vspace{1ex}
\bibliography{refs}
\bibliographystyle{plainnat}

\appendix
\numberwithin{equation}{section}
\newpage
    
  \hrule height 4 pt
  \vskip 0.25in
  \vskip -\parskip%
{\centering
{\LARGE\bf Supplementary Material:\\Graph Neural Network Bandits\par }}
  \vskip 0.29in
  \vskip -\parskip
  \hrule height 1 pt
  \vskip 0.09in%

\section{The Neural Tangent Kernel and its Connection to the GNTK}\label{app:ntk_gntk}
Let $f(\vx; \vtheta): \mathbb{R}^d \rightarrow \mathbb{R}$ be a fully-connected network, with $L$ hidden layers of equal width $m$, and ReLU activations, recursively defined as follows:
\begin{equation}
    \label{eq:def_nn}
    \begin{split}
        f^{(1)}(\vx) &= \mW^{(1)}\vx,\\
     f^{(l)}(\vx) &= \sqrt{\frac{2}{m}} \mW^{(l)} \sigma_{\text{relu}}\big(f^{(l-1)}(\vx)\big) \in \mathbb{R}^m, \,\, 1<l\leq L\\
    \nn(\vx; \vtheta) &= \sqrt{2} \mW^{(L+1)} \sigma_{\text{relu}}\big(f^{(L)}(\vx)\big)\in \sR.  
    \end{split}
\end{equation}
The weights $\mW^{(i)}$ are initialized to random matrices with standard normal i.i.d.~entries, and $\vtheta^0 = (\mW^{(i)})_{i\leq L+1}$. 
Consider the first order approximation of $\nn(\vx, \vtheta)$ around the initial parameters $\vtheta^0$, i.e., 
\begin{align*}
    \tilde f_{\mathrm{NN}}(\vx; \vtheta) &= \gradnn^T(\vx, \vtheta^0)(\vtheta - \vtheta^0),
\end{align*}
since the network is defined to be zero at initialization. By considering a fixed dataset and a square loss, training with the linear model $\tilde f_{\mathrm{NN}}(\vx, \vtheta)$, is equivalent to regression with the {\em tangent kernel} \citep{jacot2018neural}, defined as 
\begin{align}
   \tilde k_{\mathrm{NN}} (\vx, \vx') & = \gradnn^T(\vx; \vtheta^0)\gradnn(\vx'; \vtheta^0).  \label{eq:finite_ntk}
\end{align}
The tangent kernel is random since it depends on $\vtheta^0$. 
 \citet{jacot2018neural} show that in the infinite width limit, $\tilde k_{\mathrm{NN}} (G, G')$ converges to a deterministic kernel, which they call the Neural Tangent Kernel (NTK),
\[
\lim_{m \rightarrow \infty} \tilde k_{\mathrm{NN}} (\vx, \vx')/m = \ntk(\vx, \vx').
\]
The NTK satisfies the Mercer condition and has the following Mercer decomposition \citep{bietti2020deep},
\begin{equation}\label{eq:mercer_dec_ntk}
    \ntk(\vx, \vx') = \sum_{r=0}^\infty \lambda_r \sum_{s=1}^{M(d,r)} Y_{s,r}(\vx)Y_{s,r}(\vx'),
\end{equation} 
where $\{Y_{s,r}\}_{s \leq M(d,r)}$ form an orthonormal basis for $V(d,r)$ the space of degree-$r$ polynomials on $\sS^{d-1}$. They eigenvalues $\lambda_r$ decay at a $r^{-d}$ rate \citep{bietti2020deep}. The eigenfunction $Y_{s,r}$ is the $s$-th spherical harmonic polynomial of degree $r$, and $M(d,r) = \text{dim}(V_{d,r})$ gives the total count of such polynomials, where
\[M(d, r) = \frac{2k+d-2}{r} \binom{r+d-3}{d-2}.\]
The NTK adopts a recursive definition (see \cref{app:ntk_formula} ). Its properties and connections to infinite-width fully-connected networks are studied in detail \citep{arora2019exact, bietti2020deep,cao2019generalization}. 

\subsection{Properties of GNN and GNTK} \label{app:gnn_gntk_props}
We first note the connection between $\gnn$ and $\nn$.
\begin{lemma}[GNN as sums of NNs]\label{lem:main_nn_gnn} Consider $\gnn$ the graph neural network defined in \cref{eq:def_gnn}, and the feedforward network $\nn$ as given in \cref{eq:def_nn}. Then,
\[
\gnn(G, \vtheta) = \frac{1}{N}\sum_{j=1}^N \nn(\bar\vh_{G, j}, \vtheta).
\]
\end{lemma}
\begin{proof}[\textbf{Proof of \cref{lem:main_nn_gnn}}]
According to \cref{eq:def_nn}, the two layer NN with width $m$ decomposes as: 
\begin{equation} \label{eq:two_layer_nn}
    \nn(\vx; \vtheta) = \sqrt{2} \sum_{j=1}^m w_j^{(2)} \sigma_{\text{relu}}\big(
    \langle \vw_j^{(1)}, \vx \rangle \big),
\end{equation}
where $\vw_j^{(1)} \in \sR^d$ are weights in the first layer and $w_j^{(2)} \in \mathbb{R}$ are weights in the second layer. Similarly, the two layer GNN (see \cref{eq:def_gnn}) is given by:
\begin{align}
    \gnn(G; \vtheta) &= \frac{\sqrt{2}}{N} \sum_{i = 1}^N \sum_{j=1}^m w_j^{(2)} \sigma_{\text{relu}}\big(
    \langle \vw_j^{(1)}, \bar{\vh}_{G, i} \rangle \big) \\
    &= \frac{1}{N} \sum_{i = 1}^N \nn(\bar{\vh}_{G, i}; \vtheta), \label{eq:via_nn_def}
\end{align}
where \cref{eq:via_nn_def} follows from \cref{eq:two_layer_nn}. The relation in \cref{eq:via_nn_def} holds trivially for arbitrary $L$.\looseness=-1
\end{proof}
We are now ready to show the permutation invariance property.
\begin{lemma}[Geometric Invariance of $\gnn$]\label{prop:gnn_invariance}
The graph neural network $\gnn$ is invariant to node permutations, i.e., for all $G \in \cG$ and $c \in P_N$,
\[\gnn(G; \vtheta) = \gnn(c\cdot G; \vtheta)\]
\end{lemma}
\begin{proof}[\textbf{Proof of \cref{prop:gnn_invariance}}] \label{proof:gnn_invariance}
Consider any permutation $c \in P_N$. By \cref{lem:main_nn_gnn},
\begin{equation} \label{eq:gnn_perm_invar}
    \gnn(c \cdot G; \vtheta) =  \frac{1}{N}\sum_{j=1}^N \nn(\bar{\vh}_{G,c(j)}; \vtheta) = \frac{1}{N}\sum_{i=1}^N \nn(\bar{\vh}_{G,i}; \vtheta)  = \gnn(G; \vtheta).
\end{equation}
Since the summation over $\nn(\bar{\vh}_{G,c(j)}; \vtheta)$ for all $j$, contains the same terms as a sum over $\nn(\bar{\vh}_{G,i}; \vtheta)$ for all $i$.
\end{proof}

We now prove that $\gntk$ as defined in \cref{sec:problem_statement}, is deterministic and can be written as a double sum of $\ntk$'s evaluated on $\bar \vh_{G,j}$ aggregated features of different nodes of the graph.
\begin{proof}[\textbf{Proof of \cref{lem:main_gntk_ntk}}] 
For any $\vtheta^0$ we first show that,
\begin{equation}\label{eq:finit_gntk_ntk}
\tilde k_{\mathrm{GNN}} (G, G') = \frac{1}{N^2} \sum_{j, j'=1}^N \tilde k_{\mathrm{NN}} (\bar \vh_{G, j},\bar \vh_{G', j'}),
\end{equation}
where $\tilde{k}_{\mathrm{NN}}(\cdot,\cdot)$ is from \cref{eq:finite_ntk}.
Then, we take the $m \rightarrow \infty$ limit.
Starting from the definition of $\tilde k_{\mathrm{GNN}}$ (see \cref{eq:finite_gntk}) and by omitting $\vtheta^0$ for simplicity of notation, we have:
\begin{align*}
    \tilde k_{\mathrm{GNN}} (G, G') & =  \ggnn^T(G)\ggnn(G')\\
    & \stackrel{\text{\cref{lem:main_nn_gnn}}}{=}  \left[ \sum_{j=1}^N \frac{1}{N}\gradnn^T(\bar\vh_{G,j})\right]
    \left[ \sum_{j'=1}^N \frac{1}{N} \gradnn(\bar\vh_{G',j'}) \right]\\
    & = \frac{1}{N^2}  \sum_{j, j'=1}^N \gradnn^T(\bar\vh_{G,j'})
    \gradnn^{(j)}(\bar\vh_{G',j})\\
    & \stackrel{\text{\ref{eq:finite_ntk}}}{=} \frac{1}{N^2}  \sum_{j,j'=1}^N \tilde k_{\mathrm{NN}} (\bar \vh_{G, j},\bar \vh_{G', j'}).
\end{align*}
 The chain of equations above prove \cref{eq:finit_gntk_ntk}. 
Plugging in the definition of the GNTK, we obtain:
\begin{align*}
    \gntk(G, G') & =\lim_{m \rightarrow \infty} \tilde k_{\mathrm{GNN}} (G, G') /m\\
    & = \lim_{m \rightarrow \infty} \frac{1}{N^2} \sum_{j,j'=1}^N \tilde k_{\mathrm{NN}} (\bar \vh_{G, j},\bar \vh_{G', j'})/m\\
    & = \frac{1}{N^2} \sum_{j,j'=1}^N \lim_{m \rightarrow \infty} \tilde k_{\mathrm{NN}} (\bar \vh_{G, j},\bar \vh_{G', j'})/m \\
    & = \frac{1}{N^2} \sum_{j,j'=1}^N \ntk (\bar \vh_{G, j},\bar \vh_{G', j'}),
\end{align*}
where the second equality holds since $\tilde k_{\mathrm{NN}}$ is continuous, and for continues functions, limit of finite sums is equal to sum of the limits. This concludes the proof.
\end{proof}

\begin{proof}[\textbf{Proof of \cref{lem:main_gntk_perminvar}}]
From \cref{lem:main_gntk_ntk}, we have
\[
\gntk(G, G') = \frac{1}{N^2}\sum_{j,j'=1}^N \ntk(\bar \vh_{G,j}, \bar \vh_{G, j'}).
\]
It then suffices to show that $\bar k(G,G')$ (as defined in \cref{eq:perm_invar_def}) is equal to the right hand side of the above equation. 
Consider $P_N$ the set of permutations of $N$ elements. Every permutation $c \in P_N$ gives a mapping from $(1, \cdots, j, \cdots, N)$ to $(c(1), \cdots, c(j), \cdots, c(N))$, where $c(j) \in [N]$ denotes the element that is placed at the $j$-th position.
We define a restricted set of permutations 
$P_{N \vert j \rightarrow i} = \{c \in P_N :  c(j) = i\}$, such that 
\begin{equation}\label{eq:restricted_perm}
    P_N = \bigcup_{i=1}^N P_{N \vert j \rightarrow i}. 
\end{equation}
Moreover, for any $1\leq j\leq N$, $\lbrace P_{N \vert j \rightarrow i} \rbrace_{i=1}^N$ are disjoint sets and the cardinality of each restricted permutation set is
\begin{equation}\label{eq:restricted_perm_size}
\left\vert P_{N \vert j \rightarrow i}\right \vert = (N-1)!,
\end{equation}
which implies that the mapping $j \rightarrow i$ is repeated $(N-1)!$ times across the elements of $P_N$.
Back to definition of $\bar k(G, G')$ we may decompose $P_N$ and write
\begin{align*}
\bar k(G,G') & = \frac{1}{N!} \sum_{c' \in P_N} \left[ \frac{1}{N!} \sum_{c \in P_N} \frac{1}{N} \sum_{j = 1}^N \ntk(\bar\vh_{G,c(j)}, \bar \vh_{G',c'(j)}) \right]\\
& = \frac{1}{N!} \sum_{c' \in P_N} \left[ \frac{1}{N!} \frac{1}{N} \sum_{j = 1}^N \sum_{c \in P_N} \ntk(\bar\vh_{G,c(j)}, \bar \vh_{G',c'(j)}) \right] \\
& \stackrel{\text{\cref{eq:restricted_perm}}}{=} \frac{1}{N!} \sum_{c' \in P_N} \left[ \frac{1}{N!} \frac{1}{N} \sum_{j = 1}^N  \sum_{i=1}^N \sum_{c \in P_{N\vert j \rightarrow i}} \ntk(\bar\vh_{G,c(j)}, \bar \vh_{G',c'(j)})\right].
\end{align*}
Now, by definition of $P_{N \vert i}$, we have $c(j) = i$ for all $c$ in this set. Therefore,
\begin{align*}
\bar k(G,G') & =  \frac{1}{N!} \sum_{c' \in P_N} \left[ \frac{1}{N!} \frac{1}{N} \sum_{j = 1}^N  \sum_{i=1}^N \sum_{c \in P_{N\vert j \rightarrow i}} \ntk(\bar\vh_{G,i}, \bar \vh_{G',c'(j)})\right]\\
& \stackrel{\text{\cref{eq:restricted_perm_size}}}{=}   \frac{1}{N!} \sum_{c' \in P_N}  \left[ \frac{1}{N!} \frac{1}{N} \sum_{j = 1}^N  (N-1)! \sum_{i=1}^N  \ntk(\bar\vh_{G,i}, \bar \vh_{G',c'j)})\right]\\
& =  \frac{1}{N!} \sum_{c' \in P_N} \left[ \frac{1}{N^2} \sum_{j = 1}^N  \sum_{i=1}^N \ntk(\bar\vh_{G,i}, \bar \vh_{G',c'(j)})\right].
\end{align*}
Now, we consider the restricted permutations $P_N \vert j \rightarrow i'$ and repeat a similar treatment for $c' \in P_N$,
\begin{align*}
    \bar k(G, G') & =  \frac{1}{N^2} \sum_{j = 1}^N \frac{1}{N!} \sum_{c' \in P_N}  \sum_{i=1}^N \ntk(\bar\vh_{G,i}, \bar \vh_{G',c'(j)})\\
    & = \frac{1}{N^2}  \sum_{j = 1}^N  \frac{1}{N!} \sum_{i'=1}^N \sum_{c' \in P_N \vert j \rightarrow i'}  \sum_{i=1}^N \ntk(\bar\vh_{G,i}, \bar \vh_{G',c'(j)})\\
    & = \frac{1}{N^2}  \sum_{j = 1}^N \frac{1}{N!} \sum_{i'=1}^N \sum_{c' \in P_N \vert j \rightarrow i'}   \sum_{i=1}^N \ntk(\bar\vh_{G,i}, \bar \vh_{G',i'}) \\
    & =   \frac{1}{N^2}  \sum_{j = 1}^N \frac{1}{N!} \sum_{i'=1}^N (N-1)!  \sum_{i=1}^N \ntk(\bar\vh_{G,i}, \bar \vh_{G',i'})\\
    & = \frac{1}{N^2} \frac{1}{N!} \sum_{i'=1}^N N!  \sum_{i=1}^N \ntk(\bar\vh_{G,i}, \bar \vh_{G',i'})\\
    & = \frac{1}{N^2}  \sum_{i'=1}^N \sum_{i=1}^N \ntk(\bar\vh_{G,i}, \bar \vh_{G',i'}).
\end{align*}
\end{proof}

\begin{lemma}[Mercer Decomposition of the GNTK] \label{lem:gntk_mercer}
 The GNTK is Mercer and can be decomposed as \looseness -1
 \begin{equation*}
    \gntk (G, G') =  \sum_{r=0}^\infty \lambda_r\sum_{s=1}^{M(d,r)} Z_{ s,r}(\bar \vh_G) Z_{ s,r}(\bar \vh_{G'})
\end{equation*}
where $\lambda_{k}$ are identical to eigenvalues of $\ntk$. The algebraic multiplicity of each $\lambda_{r}$ is $ M(d,r)$. The eigenfunctions $\{Z_{s,r}\}_{s \leq M(d,r)}$ are degree-$r$ polynomials with the permutation invariant additive structure
\[
Z_{s,r}(\bar \vh_G) := \frac{1}{N}\sum_{j=1}^N Y_{s,r}(\bar \vh_{G,j}).
\]
where $Y_{s,r}$ are degree-$r$ spherical harmonics.
\end{lemma}
\begin{proof}[\textbf{Proof of \cref{lem:gntk_mercer}}]
Plugging in the Mercer decomposition of $\ntk$ as given in \cref{eq:mercer_dec_ntk} into \cref{lem:main_gntk_ntk} we get,
\begin{equation}\label{eq:mercer_dec_gntk}
\begin{split}
        \gntk(G, G')  & =  \frac{1}{N^2}\sum_{j,j'=1}^N \sum_{r=0}^\infty \lambda_r \sum_{s=1}^{ M(d,r)} Y_{s,r}(\bar  \vh_{G,j}) Y_{s,r}(\bar \vh_{G', j'}) \\
        & =  \sum_{r=0}^\infty \lambda_r \sum_{s=1}^{ M(d,r)} \left(  \frac{1}{N}\sum_{j=1}^N Y_{s,r}(\bar  \vh_{G,j})\right) \left( \frac{1}{N}\sum_{j'=1}^N Y_{s,r}(\bar \vh_{G', j'})\right)\\
        & = \sum_{r=0}^\infty \lambda_r \sum_{s=1}^{ M(d,r)} Z_{s,r}(\bar\vh_{G})Z_{s,r}(\bar\vh_{G'}).
\end{split}
\end{equation}

\end{proof}

\subsection{Recursive Expression for the NTK} \label{app:ntk_formula}
For the sake of completeness, we provide a closed-form expression for the NTK function used in \cref{eq:gntk_vs_ntk} (for more details, see Section 2.1 in \cite{bietti2020deep}). We limit the input space to $\mathbb{S}^{d-1}$ since, by the definition, our feature vectors are always normalized, i.e., $\|\bar{\vh}_u\|_2=1$ for every $u \in V(G)$. For a ReLU network with $L$ layers considered in \cref{eq:def_nn} with inputs on the sphere (by taking appropriate limits on the widths), the corresponding $\ntk(\vx, \vx')$ (\cite{jacot2018neural}) depends on $\angle (\vx,\vx')$ and is given by $\ntk(\vx, \vx') = \kappa^{(L)}_{\text{NN}}(\vx^T \vx')$ where
$\kappa^{(L)}_{\text{NN}}(\cdot)$ is defined recursively as follows:\looseness=-1

\begin{equation}
    \label{eq:ntk_formula}
\begin{split}
        \kappa_{\text{NN}}^{(1)}(u)&= \kappa^{(1)}(u) = u,  \\
        \kappa^{(l)}(u) & =  \kappa_1\big(\kappa^{(l-1)}(u)\big), \\
        \kappa_{\text{NN}}^{(l)}(u) & = \kappa_{\text{NN}}^{(l-1)}(u)\cdot \kappa_0\big(\kappa^{(l-1)}(u)\big) + \kappa^{(l)}(u) \quad \text{for } 2 \leq l \leq L,
        \end{split}
\end{equation}
where
\begin{align*}
\kappa_0(u) & = \frac{1}{\pi}\big(\pi-\arccos (u)\big),\\
\kappa_1(u) & = \frac{1}{\pi}\Big( u(\pi - \arccos(u)) + \sqrt{1-u^2} \Big).
\end{align*}
Finally, we note that $\kappa^{(L)}_{\text{NN}}(1) = 1$ (\citet{bietti2020deep}), and hence $\ntk(\vx, \vx') \leq 1$ for all $\vx,\vx' \in \mathbb{S}^{d-1}$.\looseness-1

\subsection{Effect of Structure on the Hypothesis Space}\label{app:shrink}

In \cref{sec:gnnucb}, we demonstrated that the additive permutation invariant structure of $\gntk$ help produce tighter bandit regret and information gain bounds, when the reward function is also permutation invariant.
We now characterize how this invariance alters the hypothesis space, independent of the bandit setup.
 \citet{bietti2020deep} give the Mercer decomposition of an NTK defined on a $Nd$-dimensional sphere. Applying their result we may decompose $\ntklong$ as
\[
 \ntklong(G, G') = \sum_{r\geq 0}^\infty \lambda_{\mathrm{NN}, r} \sum_{s=1}^{N(Nd, r)} Y_{s,r,Nd}(\bar \vh_G) Y_{s,r,Nd}(\bar \vh_{G'})
\]
where $Y_{s,r}: \sS^{Nd-1}\rightarrow \sR$ is the $s$-th degree-$r$ spherical harmonic polynomial. Each eigenvalue $ \lambda_{\mathrm{NN}, r}$ corresponds to the eigenspace $V_{Nd,r}$, the space of degree-$r$ spherical harmonics, defined on an $Nd$-dimensional domain. The algebraic multiplicity of each eigenvalue is $M(Nd,r)$ and equal to the dimension of its eigenspace, 
\[\dim(V_{Nd,r}) = M(Nd,r) = \frac{2k+d-2}{r} \binom{r+d-3}{d-2} = cr^{Nd-2}.\]
Lastly, the eigenvalues decay at a $\lambda_{\mathrm{NN}, r} \simeq c(Nd,L) r^{-Nd}$ rate. The dependence of $c(Nd,L)$ on $Nd$ is exponential, but linear in $L$, as shown by \cite{bietti2020deep}.

\begin{figure}
    \centering 
    \includegraphics[width = 0.6\linewidth]{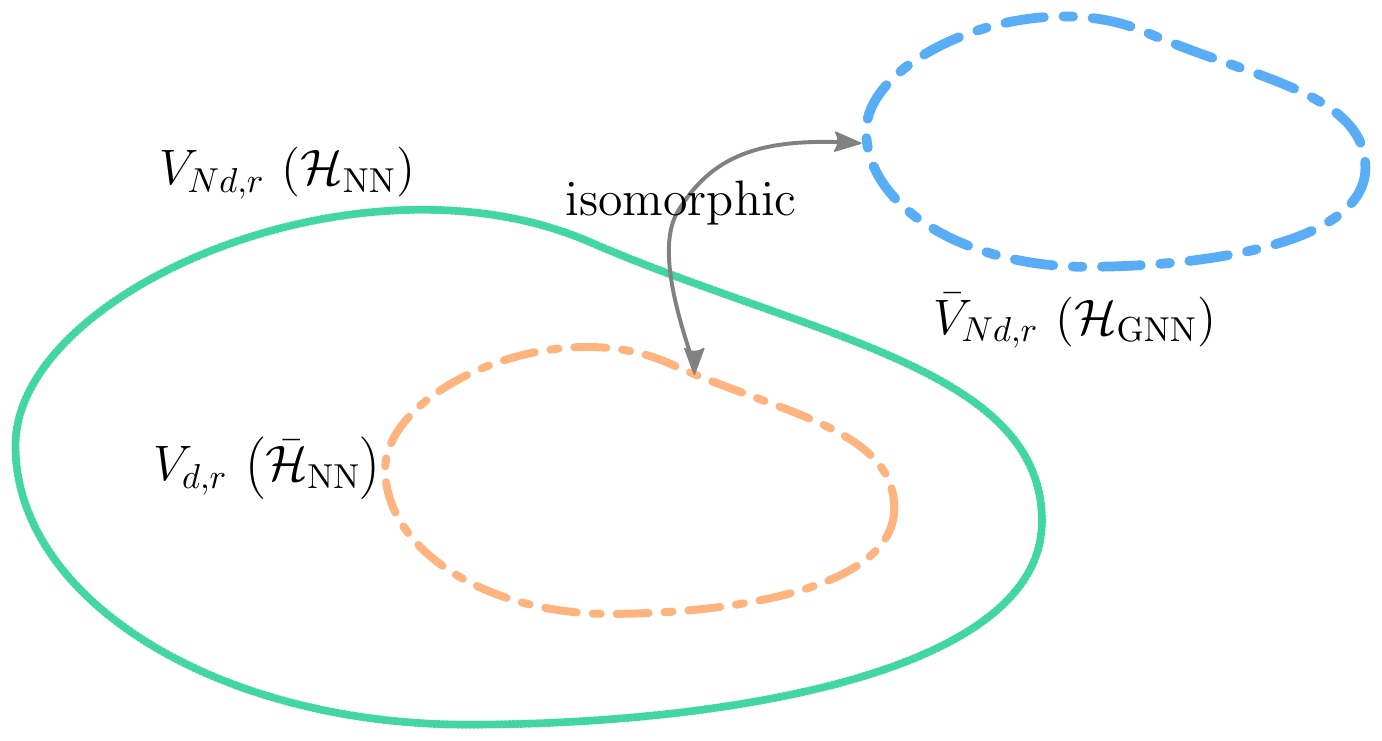}
    \caption{\label{fig:RKHS_shrink} Relation between the GNN and NN hypothesis space}
\end{figure}

We compare this kernel to the GNTK, which has the invariances encoded in its construction. In \cref{lem:gntk_mercer} we show that it may be written as 
\[
 \gntk(G, G') = \sum_{r\geq 0}^\infty \lambda_{\mathrm{GNN},r} \sum_{s=1}^{N(d, r)} Z_{s,r,Nd}(\bar \vh_G) Z_{s,r,Nd}(\bar \vh_{G'}),
\]
where the eigenvalues $\lambda_{\mathrm{GNN},r}$ decay at a $c(d,L)r^{-d}$ rate, and have an algebraic multiplicity of $M(d,r) \simeq c r^{d-2}$. The eigenvectors $Z_{s,r}$ are degree-$r$ polynomials with the permutation invariant additive structure
\[
Z_{s,r,Nd}(\bar \vh_G) := \frac{1}{N}\sum_{j=1}^N Y_{s,r,d}(\bar \vh_{G,j}).
\]
where $Y_{s,r,d}$ are degree-$r$ spherical harmonics, defined on $\sS^{d-1}$. Let $\bar V_{Nd,r}$ be the eigenspace corresponding to $\lambda_{\mathrm{GNN}, r}$ the $r$-th eigenvalue. Due to the specific structure of the $Z_{s,r, Nd}$ polynomials, there exists a bijection between $\bar V_{Nd,r}$ and $V_{d,r}$ the space of degree-$r$ spherical harmonics defined on $\sS^{d-1}$. The two vector spaces are isomorphic and thus have the same (finite) dimensionality, $\dim(\bar V_{Nd,r}) = \dim(V_{d,r})$.
This implies that the $r$-th eigenspace of the GNTK is smaller than the $r$-th eigenspace of the NTK
\[
\frac{\dim(\bar V_{Nd,r})}{\dim(V_{Nd,r})} =  \frac{\dim( V_{d,r})}{\dim(V_{Nd,r})} \leq \frac{r^{d}}{r^{Nd}} = \frac{1}{r^{d(N-1)}}.
\]
Further, note that $V_{d,r} \subsetneq V_{Nd,r}$. This connection is shown in Figure \ref{fig:RKHS_shrink}.

                  \section{Information Gain Bounds} \label{app:infogain}
In this section, we present a bound for the maximum information gain (defined in \cref{eq:MMI_def})  of the graph neural tangent kernel using the fact that a GNTK can be decomposed as an average of lower-dimensional NTKs (see \cref{eq:gntk_vs_ntk}). 
\begin{proof}[\textbf{Proof of Theorem \ref{thm:MIG_gntk}}]
We follow a similar technique as in \citet{vakili2020information}.
Consider an arbitrary sequence of graphs $(G_i)_{i\leq T}$, where each $G_i \in \cG$. Let $\mgntk \in \sR^{T\times T}$ denote the corresponding GNTK matrix where 
\[
[\mgntk]_{i,j} = \gntk(G_i, G_j) = \frac{1}{N^2}\sum_{u,u'=1}^N \sum_{r=0}^\infty \lambda_r \sum_{s=1}^{ M(d,r)} Y_{s,r}(\bar  \vh_{G_i,u}) Y_{s,r}(\bar \vh_{G_j, u'}),
\]
holds by \cref{lem:gntk_mercer}.
We decompose $\gntk(G_i, G_j)$ into $k_D(G_i, G_j)+ k_O(G_i, G_j)$ where 
\begin{align*}
    k_D (G_i, G_j) & =  \frac{1}{N^2}\sum_{u,u'=1}^N \sum_{r\leq D}\lambda_r \sum_{s=1}^{M(d,r)} Y_{s,r}(\bar \vh_{G_i,u})Y_{s,r}(\bar \vh_{G_j, u'}),\\
    k_O (G_j, G_j) & = \frac{1}{N^2}\sum_{u,u'=1}^N \sum_{r\geq D+1}\lambda_r \sum_{s=1}^{M(d,r)} Y_{s,r}(\bar \vh_{G_i,u})Y_{s,r}(\bar \vh_{G_j, u'}).
\end{align*}
The kernel $k_D$ is reproducing for $\cH_{k_D}$, a finite-dimensional subspace of $\cH_\gntk$ that is spanned by the eigenfunctions corresponding to the first $D$ \textit{distinct} eigenvalues. The kernel $K_O$ is reproducing for $\cH_{k_O}$ which is orthogonal to $\cH_{k_D}$. Moreover 
$
k_D(G_i,G_j)= \vphi_D^T(G_i)\vphi_D(G_j)
$
where the concatenated feature vector $\vphi_D(G)$ is defined as
\[
\vphi_D(G)=\left(\left(\frac{\sqrt{\lambda_0}}{N}\sum_{u=1}^N Y_{s,0}(\bar \vh_{G,u}),\right)_{s\leq N(d,0)}, \cdots, \left(\frac{\sqrt{\lambda_D}}{N}\sum_{u=1}^N Y_{s, D}(\bar \vh_{G,u})\right)_{s\leq N(d,D)}\right).
\]
Here, $\vphi_D (G) \in \sR^{\tilde D}$ where
\begin{equation}\label{eq:tilde_D}
\tilde D = \sum_{r=0}^{D} M(d,r) \simeq C  \sum_{r=0}^{D} r^{d-2} \leq C  \frac{(D+1)^{d-1}}{D-1},
\end{equation}
since by Stirling's approximation $M(d,r)$ grows with $r^{d-2}$ \citep{bietti2020deep}. 
 
\looseness=-1 Recall that the information gain is $I(\vy_T; \vf_T) = \frac{1}{2}\log\det(\mI + \lambda^{-1}\mgntk)$. 
Defining $\mK_D$ and $\mK_O$ such that $[\mK_D]_{i,j} = k_D(G_i, G_j)$ and $[\mK_O]_{i,j}= k_O(G_i,G_j)$, we get $\mgntk = \mK_D + \mK_O$ and therefore,
\begin{equation} \label{eq:infogain_decomp}
    \begin{split}
        I(\vy_T; \vf_T) & = \frac{1}{2}\log\det(\mI + \lambda^{-1}(\mK_D+\mK_O))\\
        & =  \frac{1}{2}\log\det(\mI + \lambda^{-1}\mK_D) + \frac{1}{2}\log\det(\mI + (\mI + \lambda^{-1}\mK_D)^{-1}\mK_O). \\
    \end{split}
\end{equation}
We bound each term separately, starting with the first term.

Consider the $T\times \tilde D$ feature matrix $\mPhi_D = \left[\vphi_D(G_1), \cdots, \vphi_D(G_T) \right]$.
Then $\mK_D = \mPhi_D \mPhi_D^T$ and by the Weinstein-Aronszajn identity,
\[
\frac{1}{2}\log\det(\mI + \lambda^{-1}\mK_D) = \frac{1}{2}\log\det(\mI + \lambda^{-1}\mPhi^T_D \mPhi_D).
\]
For positive definite matrices $\mP \in \mathbb{R}^{n\times n}$, we have $\log\det \mP \leq n \log \tr (\mP/n)$. Applying this identity we get,
\begin{equation}\label{eq:gamma_firstterm}
    \begin{split}
        \frac{1}{2}\log\det(\mI + \lambda^{-1}\mK_D) & \leq \frac{1}{2} \tilde D \log\left(  1 +  \frac{\lambda^{-1}}{\tilde D}\tr\left(\mPhi^T_D \mPhi_D \right) \right) \\
    & =\frac{1}{2} \tilde D \log\left(  1 +  \frac{\lambda^{-1}}{\tilde D}\sum_{t=1}^T \vphi^T_D(G_t) \vphi_D(G_t) \right) \\
    & =\frac{1}{2} \tilde D \log\left(  1 +  \frac{\lambda^{-1}}{\tilde D}\sum_{t=1}^T \vert\vert\vphi_D({G_t})\vert\vert_2^2 \right) \\ 
    &  \leq \frac{1}{2} \tilde D \log\left(  1 +  \frac{\lambda^{-1}}{\tilde D}\sum_{t=1}^T \sum_{r=0}^{D} \frac{\lambda_r}{N^2}\sum_{s=1}^{M(d,r)} \left(\sum_{u = 1}^N Y_{s,r}(\bar \vh_{G_t, u}) \right)^2\right) \\ 
    & \leq \frac{1}{2} \tilde D \log\left(  1 +  \frac{\lambda^{-1}}{\tilde D}\sum_{t=1}^T \sum_{u, u' = 1}^N \sum_{r=0}^{D} \frac{\lambda_r}{N^2}\sum_{s=1}^{M(d,r)} Y_{s,r}(\bar \vh_{G_t,u}) Y_{s,r}(\bar \vh_{G_t,u'}) \right) \\ 
    &  = \frac{1}{2} \tilde D \log\left(  1 +  \frac{\lambda^{-1}}{\tilde D}\sum_{t=1}^T k_D(G_t, G_t) \right) \\ 
    & \leq \frac{1}{2} \tilde D \log\left(  1 +  \frac{T/\lambda}{\tilde D}\right).
    \end{split}
\end{equation}
The last inequality holds since by definition, $\gntk$ is uniformly bounded by $1$ on the unit sphere (this holds since $\ntk$ is also uniformly bounded by $1$ on the same domain; see \cref{app:ntk_formula}). 

For bounding the second term, we again use the  $\log\det \mP \leq n \log \tr (\mP/n)$ inequality and write,
\begin{equation*} 
\begin{split}
        \frac{1}{2}\log\det(\mI + (\mI + \lambda^{-1}\mK_D)^{-1}\mK_O) & \leq \frac{T}{2}\log\left( 1 + \frac{\tr\left((\mI + \lambda^{-1}\mK_D)^{-1}\mK_O\right)}{T}\right) \\
        &\leq
        \frac{T}{2}\log\left( 1 + \tr\left(\mK_O\right)/T\right).
\end{split}
\end{equation*}
The second inequality holds due to $(\mI + \lambda^{-1}\mK_D)^{-1}$ being positive definite, with eigenvalues smaller than $1$. To bound $\tr(\mK_O)$, note that
\begin{align*}
    [\mK_O]_{i,j} & = \frac{1}{N^2} \sum_{u,u'=1}^N \sum_{r\geq D+1}\lambda_r \sum_{s=1}^{M(d,r)} Y_{s,r}(\bar \vh_{G_i,u})Y_{s,r}(\bar \vh_{G_j,u'})\\
    & = \frac{1}{N^2} \sum_{u,u'=1}^N \sum_{r\geq D+1}\lambda_r M(d,r) P_r(\langle \bar \vh_{G_i,u}, \bar \vh_{G_j,u'} \rangle )\\
    & \leq \frac{1}{N^2} \sum_{u,u'=1}^N \sum_{r\geq D+1}\lambda_r M(d,r)\\
    & \leq \sum_{r\geq D+1}\lambda_r M(d,r)
\end{align*}
The second equality follows from $\sum_{s=1}^{M(d,r)} Y_{s,r}(\vx)Y_{s,r}(\vx') = M(d,r) P_r(\vx^T \vx')$, where $P_r$ is the degree-$r$ Legendre polynomial \citep{bietti2020deep}. The Legendre basis is bounded in $[0,1]$, resulting in the first inequality.
By \citet[][Corollary 3]{bietti2020deep}, there exists a constant $c_1(d, L)$ such that $\lambda_r \leq c_1(d, L) r^{-d}$. Stirling's approximation states that $n! \sim \sqrt{2\pi n}(n/e)^n$. Therefore, there exists a constant $c_2$ such that $M(d,r) \leq c_2 r^{d-2}$. Therefore, there exists a constant $c(d, L)$ such that,
\begin{align*}
    [\mK_O]_{i,j}  \leq  \sum_{r\geq D+1}\lambda_r M(d,r)  \leq c(d, L) \sum_{r\geq D+1} r^{-2} \leq \frac{c(d, L)}{D}
\end{align*}
where the second inequality comes from
\[
\sum_{r\geq D+1} r^{-2} \leq \int_D^\infty z^{-2}\mathrm{d}z = \frac{1}{D}.
\]
Therefore, we may bound the second term of the information gain as follows,
\begin{equation}\label{eq:gamma_secondterm}
\frac{1}{2}\log\det(\mI + (\mI + \lambda^{-1}\mK_D)^{-1}\mK_O) \leq  \frac{T}{2} \log\left( 1 + \frac{c(d, L)}{D} \right) 
\end{equation}
From \cref{eq:infogain_decomp}, \cref{eq:gamma_firstterm} and \cref{eq:gamma_secondterm},
\begin{equation}\label{eq:infogain_last}
    I(\vy_T; \vf_T)  \leq \frac{1}{2} \tilde D \log\left(  1 +  \frac{T\lambda^{-1}}{\tilde D}\right) + \frac{T}{2} \log\left( 1 + \frac{c(d, L)}{D} \right).
\end{equation}
For the first term to dominate the second, $\tilde D$ has to be set to
\[
\tilde D = \Big\lceil \left(\frac{c(d, L) T}{\log (1+ T/\lambda)}\right)^{\frac{d-1}{d}}\Big\rceil.
\]
This results in
\[
        \gamma_T = \mathcal{O}\left( \left( \frac{T}{\log (1+ \frac{T}{\lambda})} \right)^{\frac{d-1}{d}}
         \log\left( 1
          + \frac{ T}{\lambda}\left( \frac{\log (1+ \frac{T}{\lambda})}{T} \right)^{\frac{d-1}{d}} \right)\right),
\]
that is itself $\cO(T^{\frac{d-1}{d}}\log^{\frac{1}{d}}T)$, therefore concluding the proof.
\end{proof}

\section{Proof of \texorpdfstring{\cref{thm:CI_gnn}}{} and \texorpdfstring{\cref{thm:gnnus_regret}}{}} \label{app:gnnus_regret}

\begin{algorithm}[ht] 
    \caption{\label{alg:gnnalg} \phasedalg
    }
    \DontPrintSemicolon
    \KwInput{$m,\,J,\,\eta,\,\lambda,\,T$}
    Set episode index $e=1$, episode length $T_e=1$, and set of potentially optimal graphs $\mathcal{G}_e = \mathcal{G}$\\  
    \init{ network parameters to a random $\vtheta^0$.}{}
    \For{$t = 1, \dots, T_e$}{
    For all $G \in \cG$, calculate $\hat\sigma_{t-1}(G)$ as defined in \cref{eq:GNNposteriors}.\\
    Select 
    \begin{equation}\label{eq:exploration_policy}
        G_t = \argmax_{G \in \cG_e} \hat\sigma^2_{t-1}(G)
    \end{equation}
    }
    Receive $\lbrace y_1, \dots, y_{T_e} \rbrace$, such that 
     \[y_t = f^*(G_t) + \epsilon_t \quad \text{ for } \quad t \in \lbrace 1, \dots, T_e \rbrace\] \\
    Calculate $\vtheta^{(J)}_{e} =\text{TrainGNN}\left(m ,J, \eta, \lambda, \vtheta^0, (G_t, y_t)_{t=1}^{T_e}\right)$\\
    Use $\beta$ and $\varepsilon$ as defined in \Cref{eq:beta_episodic} and \Cref{eq:e_episodic} and update
    \begin{equation} \label{eq:potential_maximizers_def}
         \mathcal{G}_{e+1} \leftarrow \big\lbrace G \in \mathcal{G}_e: \gnn(G;\vtheta^{(J)}_{e}) + \beta \hat{\sigma}_{T_e}(G) + 2\varepsilon\geq  \max_{G \in \mathcal{G}_e} \big( \gnn(G;\vtheta^{(J)}_{e}) - \beta \hat{\sigma}_{T_e}(G)\big)  \big \rbrace,
    \end{equation} \\
    Set $T_{e+1} \leftarrow 2T_e,\; e \leftarrow e + 1$ and return to the \textbf{Initialize} step  \\
    \KwOutput{Terminate after $T$ total evaluations ($E$ total episodes) and return 
    \begin{equation}\label{eq:reporting_rule_gnnus}
        \hat{G}_T = \argmax_{G \in \cG} \gnn(G; \vtheta^{(J)}_{E-1})
    \end{equation}
    }
\end{algorithm}

\begin{algorithm}[ht] 
\DontPrintSemicolon
\KwInput{$m,\,J,\,\eta,\,\lambda,\,\vtheta^0,\,(G_i, y_i)_{i < t}$}
Define 
    $\cL(\vtheta) = \frac{1}{t}\sum_{i < t } \big(\gnn(G_i; \vtheta) - y_i\big)^2 + m \lambda \norm{\vtheta - \vtheta^0}_2^2$\\
Initialize $\vtheta^{(0)} = \vtheta^0$ \\
 \For{$j = 1, \dots, J$}
{
$\vtheta^{(j)} = \vtheta^{(j-1)} - \eta \nabla\cL(\vtheta^{(j-1)})$
}
\KwOutput{$\vtheta^{(J)}$}
\caption{\label{alg:trainnn} TrainGNN}
\end{algorithm}

\begin{proof}[\textbf{Proof of \cref{thm:gnnus_regret}}]
The proposed algorithm \gnnus (see \Cref{alg:gnnalg}) is a variant of the Phased GP Uncertainty algorithm proposed in \cite{bogunovic2021misspecified}. The following analysis closely follows the one of \cite{bogunovic2021misspecified} (but ignores misspecification), with important differences pertained to the introduction of GNN estimator and GNTK analysis. The algorithm runs in episodes of exponentially increasing length $T_e$, and maintains a set of potentially optimal graphs $\mathcal{G}_e$. To compute the set of potentially optimal graphs after every episode, it uses the confidence bounds from \Cref{thm:CI_gnn}.
The total number of episodes is denoted with $E$, and it holds that $E \leq \lceil \log_2 T \rceil$, since the length of the episode is growing exponentially.\looseness=-1

To bound the regret of \gnnus, we make use of the finite-dimensional tangent kernel. With a slightly different notation from \cref{sec:gnn}, we set $\tgntk (\cdot,\cdot) =  \ggnn^T(\cdot; \vtheta^0) \ggnn(\cdot; \vtheta^0)/m$, where $\ggnn^T(\cdot; \vtheta^0)$ denotes the gradient of the GNN at initialization.
We argued in the main text that this kernel can well approximate $\gntk$. The feature map corresponding to this kernel, $\hat \vphi(G) = \ggnn(G)/m$ can be viewed as a finite-dimenional approximation of $\vphi_{\mathrm{GNN}}$, the (infinite length) feature map of the GNTK. 

Throughout the proof, we denote the posterior mean and variance calculated via $\mathrm{GP}(0, \tgntk)$ by $\hat \mu_{t-1}$ and $\hat \sigma_{t-1}$, respectively. Recall that the posterior mean and variance function after observing the data $(G_i, y_i)_{i < t}$ is calculated via
\begin{equation}
\label{eq:GPposteriors}
\begin{split}
    \hat{\mu}_{t-1} (G; \tgntk)& = {\vkhat}_{t-1}^T(G)({\mKhat}_{t-1}+\lambda\mI)^{-1}\vy_{t-1},  \\
     \hat{\sigma}^2_{t-1}(G; \tgntk) & =  \tgntk(G,G) - {\vkhat}^T_{t-1}(G)({\mKhat}_{t-1}+\lambda\mI)^{-1}{\vkhat}_{t-1}(G),
     \end{split}
     \end{equation}
\looseness=-1 where the constant $\lambda$ is the variance proxy for observation noise. Here $\vy_{t-1} = [y_i]_{i < t}$ is the vector of observed values, $\vkhat_{t-1}(\vx) = [\tgntk(\vx, \vx_\tau)]_{i < t}$, and ${\mKhat}_{t-1} = [ \tgntk(\vx_i, \vx_j)]_{i,j < t}$ is the kernel matrix. \looseness=-1

We note that \gnnus uses $\hat \sigma_{t-1}$ as the variance estimate, while instead of $\hat{\mu}_{t-1}$, the algorithm makes use of the GNN predictions, i.e., $\gnn(G; \vtheta^{(J)}_t )$ for the center of the confidence set. 

As will become clear soon, since \gnnus uses $\hat \sigma_{t-1}$, this yields a regret bound depending on $\hat\gamma_T$, the information gain corresponding to the approximate kernel $\tgntk$. Lastly, in Lemma \ref{lem:gammahat_gamma}, for appropriately set width $m$, we bound $\hat\gamma_T$ with $\gamma_{\mathrm{GNN},T}$ (maximum information gain corresponding to the exact GNTK). We use this result in our steps bellow.  



\paragraph{Step 1 (Max variance bound)}  Consider any fixed episode $e$, and recall that $T_e$ denotes the episode length. By the exploration policy of \gnnus (\Cref{eq:exploration_policy}), at any step $t$ (within an episode) and for any graph $G \in \cG_e$, we have $\hat \sigma_{t-1}(G) \leq \hat \sigma_{t-1}(G_t)$. From \cref{eq:GPposteriors}, since the covariance matrix is positive definite, conditioning on a larger set of points reduces the posterior variance and thus $\hat \sigma_{T_e}(G) \leq \hat \sigma_{t-1}(G)$, for all $G \in \cG_e$ and $t \leq T_e$. Putting the two inequalities together, $\hat \sigma_{T_e}(G) \leq \min_{t\leq T_e} \hat{\sigma}_{t-1}(G_t)$, which gives
\[
\hat \sigma^2_{T_e}(G) \leq \frac{1}{T_e}\sum_{t=1}^{T_e} \hat \sigma^2_{t-1}(G_t).
\]
For any $s \in [0, 1/\lambda]$, it holds that
\[
s^2 \leq \frac{1}{\lambda \log (1+1/\lambda)} \log (1+s^2).
\]
For any $G_t$, we have $\hat \sigma^2_{t-1}(G_t)/\lambda \leq \tgntk (G_t, G_t)/\lambda \leq 1/\lambda$. Therefore,
\[
 \hat \sigma^2_{t-1}(G_t) \leq \frac{1}{\log (1+1/\lambda)} \log (1+\lambda^{-2}\hat \sigma^2_{t-1}(G_t)).
\]
From \citet[][Lemma 5.3]{srinivas2009gaussian}, we can conclude that, 
\[
\sum_{t=1}^{T_e} \log (1 + \lambda^{-1}\hat \sigma^2_{t-1}(G_t)) = 2 I(G_1, \cdots, G_t; \tgntk) \leq 2\hat\gamma_{T_e}, 
\]
\looseness-1 where $\hat\gamma_{T_e,}$ is the maximum information gain corresponding to this episode and the kernel $\tgntk$. This inequality allows us to bound the posterior variance at the end of episode $e$ as follows. For any $G \in \cG_e$,
\begin{equation}\label{eq:post_var_bound}
   \hat \sigma^2_{T_e}(G) \leq \frac{2\hat\gamma_{T_e}}{T_e\log(1+\lambda^{-1})}.
\end{equation}
\paragraph{Step 2 (Confidence bounds)} 
Consider an episode $e$, and let $\tilde \delta = \delta/(3 E)$, where $E\leq \log T$ is the number of episode. From \cref{thm:CI_gnn_formal}, for all $G \in \cG$, with probability at least  $1-\tilde \delta$, 
\begin{equation} \label{eq:practical_CI}
\vert \gnn(G; \vtheta^{(J)}_{e}) - f^*(G) \vert \leq \beta\hat\sigma_{T_e}(G) + \varepsilon,
\end{equation}
where for simplicity we use
\begin{align}
    \beta&:= \sqrt{2}B + \frac{\sigma}{\sqrt{\lambda}}\sqrt{2\log2\vert \cG \vert/\tilde\delta} +  \sqrt{\frac{2B}{m\eta\lambda}}\left( \sqrt{2} + (1-m\eta\lambda )^{J/2}\right), \label{eq:beta_episodic}\\
    \varepsilon&:= \tilde C L^3\left(\frac{B}{m \lambda}\right)^{2/3} \sqrt{m \log m}. \label{eq:e_episodic}
\end{align}
Applying the union bound \cref{eq:practical_CI} holds for every $G\in \cG$ and $e \in [E]$ with probability at least $1-\delta/3$. 
Finally, by using $\tilde{\delta} = \delta /3$ in \cref{lem:gammahat_gamma} and by applying the union bound, we have that both events in \cref{lem:gammahat_gamma} and in \cref{eq:practical_CI} hold jointly with probability at least $1- 2\delta/3$. In the rest of the proof, we condition on the joint event holding true. This implies that $G^* \in \mathcal{G}_e$ for every $e \in [E]$, i.e., according to the rule in \cref{eq:potential_maximizers_def}, the algorithm will not eliminate $G^*$.



\paragraph{Step 3 (Cumulative Regret)} We use $R_e$ to denote the episodic regret, and write
\begin{equation} \label{eq:cumulative_regret_first}
    R_T = \sum_{e=1}^E R_e \leq m_1 B + \sum_{e=2}^E \sum_{t=1}^{T_e} \big( f^*(G^*) - f^*(G^{(e)}_t) \big).
\end{equation}
The first inequality follows since $\gntk$ is uniformly bounded by $1$ and the RKHS norm of $f^*$ is bounded by $B$. We also add an additional superscript $e$ in $G^{(e)}_t$, to denote a graph selected in episode $e$ at time step $t$.

Consider any episode $e$, the following holds due to \cref{eq:practical_CI}:
\begin{align*}
    f^*(G^*) - f^*(G^{(e)}_t) &\leq \gnn(G^*; \vtheta^{(J)}_{e-1}) + \beta \hat{\sigma}_{T_{e-1}}(G^*) + 2\varepsilon \\ &\quad- \big(\gnn(G^{(e)}_t; \vtheta^{(J)}_{e-1}) - \beta \hat{\sigma}_{T_{e-1}}(G^{(e)}_t)\big).
\end{align*}

Moreover, we use the elimination rule in \cref{eq:potential_maximizers_def} to obtain:
\begin{align}
    f^*(G^*) - f^*(G^{(e)}_t) &\leq   
    \big(\gnn(G^*; \vtheta^{(J)}_{e-1}) - \beta \hat{\sigma}_{T_{e-1}}(G^*) - \varepsilon) \nonumber \\
    & \, + 2\beta \hat{\sigma}_{T_{e-1}}(G^*) + 4\varepsilon\nonumber \\
    & \,- \big(\gnn(G^{(e)}_t; \vtheta^{(J)}_{e-1}) + \beta \hat{\sigma}_{T_{e-1}}(G^{(e)}_t)+ \varepsilon\big) \nonumber\\
    & \, + 2\beta \hat{\sigma}_{T_{e-1}}(G^{(e)}_t)\nonumber
    \\ 
    &\leq 4 \beta_{T_{e-1} + 1} \max_{G \in \mathcal{G}_{e-1}} \hat{\sigma}_{T_{e-1} }(G) + 4 \varepsilon.\label{eq:cumulative_regret_part}
\end{align}

Next, we combine \cref{eq:cumulative_regret_part} and \cref{eq:cumulative_regret_first} and write:
\begin{align}
    R_T &\leq m_1 B + \sum_{e=2}^E \sum_{t=1}^{T_e} 4 \beta_{T_{e-1} + 1} \max_{G \in \mathcal{G}_{e-1}} \hat{\sigma}_{T_{e-1} }(G)  + 4\varepsilon\\ 
    &\leq m_1 B + \sum_{e=2}^E 4 T_e  \beta \sqrt{ \frac{2\hat\gamma_{T_{e-1}}}{T_{e-1}\log(1+\lambda^{-1})}} + 4T_e \varepsilon \label{eq:cregret_1}\\
    &= m_1 B + \sum_{e=2}^E 8   \beta \sqrt{ \frac{2T_{e-1}\hat\gamma_{T_{e-1}}}{\log(1+\lambda^{-1})}} + 4T_e \varepsilon\label{eq:cregret_2}\\
    &\leq m_1 B + \sum_{e=2}^E 8 \beta \sqrt{ \frac{2 T\hat\gamma_{T}}{\log(1+\lambda^{-1})}} + 4T_e \varepsilon \label{eq:cregret_3}\\
    &\leq m_1 B +  8 (\log(T) + 1)\Big( \beta \sqrt{ \tfrac{2 T\hat\gamma_{T}}{\log(1+\lambda^{-1})}} + 4T \varepsilon \Big) \label{eq:cregret_4}\\
    &\leq m_1 B + 8 (\log(T) + 1)\Big( \beta \sqrt{\tfrac{2 T(\gamma_{T} + \epsilon(m))}{\log(1+\lambda^{-1})}} + 4T \varepsilon \Big)\label{eq:cregret_5},
\end{align}
where \cref{eq:cregret_1} follows since $T > m_{e}$ for every $e$ and \cref{eq:post_var_bound} and \cref{eq:cregret_2} since $T_e = 2T_{e-1}$. To obtain \cref{eq:cregret_3}, we use that $T > T_{e-1}$ and $\hat{\gamma}_{T} \geq \hat{\gamma}_{T_e-1}$. Finally, \cref{eq:cregret_4} follows since the number of episodes $E \leq \lceil \log T \rceil$, and \cref{eq:cregret_5} follows from \cref{lem:gammahat_gamma}.

\paragraph{Step 4 (Putting everything together)}
Plugging in the expressions for $\beta$ and $e$ we obtain

\begin{align*}
    R_T \leq &8 \sqrt{2}(\log T + 1)\sqrt{\tfrac{2 T(\gamma_{T} + \epsilon(m))}{\log(1+\lambda^{-1})}} \left( B + \tfrac{\sigma}{\sqrt{\lambda}}\sqrt{\log \tfrac{\vert \cG\vert (\log T + 1)}{\delta}} \sqrt{\tfrac{B}{m\eta\lambda}}\left(\sqrt{2}+(1-m\eta\lambda)^{J/2} \right)\right)\\
    & \quad + B + 32 (\log T+1)\tilde{C}T L^3 \left(\frac{B}{m\lambda}\right)^{2/3}\sqrt{m\log m}
\end{align*}
Since $m = \poly(t)$ and $\eta \sim 1/m$, the last term and $\epsilon(m)$ vanishes with $T$ at a $o(1)$ rate, and the gradient descent error term becomes a constant factor.
Then we obtain with probability at least $1-\delta$
\begin{equation}
    R_T = \cO\left(\sqrt{T\gamma_{T}} \left( B + \frac{\sigma}{\sqrt{\lambda}}\sqrt{\log\frac{\vert\cG\vert \log T}{\delta}}\right)\right).
\end{equation}

 \end{proof}

\begin{lemma}[\textbf{Bounding MIG with its approximation}]\label{lem:gammahat_gamma}
Set $\delta \in (0,1)$. If $m = \poly \left(t, L, \vert \cG\vert, \lambda, B, \lambda_0^{-1}, \log( N/\delta)\right)$, then with probability at least  $1-\delta$,
\[
\hat\gamma_T \leq \gamma_T + \epsilon(m),
\]
where $\gamma_T$ is the maximum information gain of the GNTK over $\cG$ as defined in \cref{eq:MMI_def},  
and $\epsilon(m) = o(m^{-1/4})$.
\end{lemma}
\begin{proof}[\textbf{Proof of Lemma \ref{lem:gammahat_gamma}}]
The proof follows from Lemma \ref{lem:gram_conv}, by repeating the technique given in Lemma D.5, \citep{kassraie2021neural}. Here we repeat it for the sake of completeness. 

Consider an arbitrary sequence of graphs $(G_t)_{t\leq T}$. Consider the feature map $\hat \vphi(G) = \ggnn(G; \vtheta^0)/\sqrt{m}$. For the kernel $\hat{k}_{\mathrm{GNN}}$, which corresponds to this feature map, the information gain after observing $T$ samples is
\[
\hat I_T = \frac{1}{2}\log\det \big(\mI + \lambda^{-1}\mgnn_T\mgnn_T^T/m\big),
\]
where $\mgnn_T = [\bar\vg(G_t)]^T_{t\leq T}$. Let $[\mgntk]_{i,j\leq T} = \gntk(G_i, G_j)$ with $k$ the NTK function of the fully-connected $L$-layer network.
\begin{equation} \label{eq:Ihat_gamma}
    \begin{split}
        \hat I_T & = \frac{1}{2}\log\det \big(\mI + \lambda^{-1}\mgntk + \lambda^{-1}(\mgnn_T\mgnn_T^T/m-\mgntk)\big)\\
        & \stackrel{\mathrm{(a)}}{\leq} \frac{1}{2}\log\det \big(\mI + \lambda^{-1}\mgntk\big) + \langle (\mI + \lambda^{-1}\mgntk)^{-1}, \lambda^{-1}(\mgnn_T\mgnn_T^T/m-\mgntk) \rangle\\
        & \leq I_T + \lambda^{-1}\norm{(\mI + \lambda^{-1}\mgntk)^{-1}}_F \norm{\mgnn_T\mgnn_T^T/m-\mgntk}_F\\
        & \stackrel{\mathrm{(b)}}{\leq}  I_T + \lambda^{-1}\sqrt{T} \norm{\mgnn_T\mgnn_T^T/m-\mgntk}_F \\
        & \stackrel{\mathrm{(c)}}{\leq}  I_T + \lambda^{-1}T\sqrt{T}\epsilon \\
        & \stackrel{\mathrm{(d)}}{\leq}  \gamma_T + \epsilon(m).
    \end{split}
\end{equation}
Inequality (a) holds by concavity of $\log\det(\cdot)$. Inequality (b) holds since $\mI  \preccurlyeq \mI + \lambda^{-1}\mgntk$. Inequality (c) holds due to Lemma \ref{lem:gram_conv}. Finally, inequality (d) uses the polynomial choice of $m$, and requires that $m$ grows with at least $O(T^6)$. Equation \ref{eq:Ihat_gamma} holds for any arbitrary context set, thus it also holds for the sequence which maximizes the information gain.
\end{proof}

\subsection{Proof of \texorpdfstring{\cref{thm:CI_gnn}}{}}\label{app:main_CI_proof}
We first present the formal version of \cref{thm:CI_gnn}.
 \begin{theorem}[GNN Confidence Bound, Formal] \label{thm:CI_gnn_formal}
Set $\delta \in (0,1)$.
Suppose $f^* \in \cH_{\gntk}$ with a bounded norm $\norm{f^*}_{\gntk}\leq B$.
Samples of $f$ are observed with zero-mean $\sigma^2$-sub-Gaussian noise. 
Assume that the random sequences $(G_i)_{i < t}$ and $(\epsilon_i)_{i<t}$ are statistically independent. 
Set $J>1$, choose the width
$
         m =   \poly \left(t, L, \vert \cG\vert,\lambda, \lambda_0^{-1}, \log( N/\delta)\right),
$
and learning rate $\eta = C(Lm+m\lambda)^{-1}$ with some universal constant $C$. Then for all graphs $G \in \cG$, with probability of at least $1-\delta$, 
\begin{equation*}
        \vert  f^*(G) -\hat\mu_{t-1}(G)\vert \leq  \beta \hat\sigma_{t}(G) + \tilde C L^3\left(\frac{B}{m \lambda}\right)^{2/3} \sqrt{m \log m}
\end{equation*}
where
\[
\beta = \sqrt{2}B + \frac{\sigma}{\sqrt{\lambda}}\sqrt{2\log\left(2\vert \cG\vert/\delta\right)} +  \sqrt{\frac{2B}{m\eta\lambda}}\big( \sqrt{2} + (1-m\eta\lambda )^{J/2}\big) 
\]
for some constant $\bar C$.
 \end{theorem}
 
To prove the theorem, first we state the necessary lemmas. 

\begin{lemma}[\textbf{Confidence interval for $\gnn$ around $\hat\mu_{t-1}$}]\label{lem:CI_gnn_approxpostmean}
Assume history $H_t = \{(G_i, y_i)\}_{i \leq t}$ with $(G_i)_{i \leq t}$ and $(\varepsilon_i)_{i\leq t}$ statistically independent. Let $m = \poly \Big(t, L, \lambda,  \log(\vert \cG\vert N/\delta)\Big)$. There exists $C_1$, such that for any $\delta>0$, if the learning rate is picked $\eta = C_1(Lm  +m\lambda)^{-1}$, then for a graph $G \in \cG$, with probability of at least $1-\delta$, 
\[
\vert \gnn(G; \vtheta^{(J)}) - \hat\mu_{t}(G) \vert \leq \hat\sigma_{t}(G) \sqrt{\frac{2B}{m\eta\lambda}}\big( \sqrt{2} + (1-m\eta\lambda )^{J/2}\big) + \tilde C L^3\left(\frac{B}{m \lambda}\right)^{2/3} \sqrt{m \log m}
\]
for some constant $\bar C$, where $\hat\mu_{t}$ and $\hat\sigma_{t}$ are as defined in \cref{eq:GPposteriors}.
\end{lemma}

\begin{proof}[\textbf{Proof of Lemma \ref{lem:CI_gnn_approxpostmean}}] 

We define $\mZ:= \lambda\mI + \mgnn^T\mgnn/(Tm)$ and $\vb = \sum_{i\leq t}y_i\bar\vg(G_i)/(T\sqrt{m})$. Recall that $\mgnn_T = [\bar\vg(G_t)]^T_{t\leq T}$. 
Let the sequence $(\tilde \vtheta^{(j)})_{j=1}^J$ denote the gradient descent updates on the following loss function
\[
\tilde\cL(\vtheta) =  \frac{1}{t}\sum_{i\leq t} \big(\langle\bar \vg(G_i),\vtheta-\vtheta^0\rangle - y_i\big)_2^2     + m \lambda \norm{\vtheta-\vtheta^0}_2^2.
\]

Note that $\tilde\vtheta^0 = \vtheta^0$ and $\tilde \vtheta^{(j)}$ also depends on $t$ the number of data points. We omit the $t$ index, for simplicity of the notation during the proof of this lemma.

By Lemma \ref{lem:GD_norm_bounds}, $\norm{\mgnn}_F \leq C\sqrt{TLm}$ and we have,
\begin{equation} \label{eq:Khat_upper}
    \mZ \stackrel{\mathrm{w.h.p}}{\preccurlyeq} (\lambda + CL)\mI \preccurlyeq \frac{1}{m\eta}\mI,
\end{equation}
since $\eta$ is set such that $\eta \leq C(m\lambda+Lm)^{-1}$. Therefore, for any $\vx\in \R^p$, $\norm{\vx}_{\mZ} \leq \frac{1}{\sqrt{m\eta}}\norm{\vx}_2$. Now consider $\vx, \vx' \in \sR^p$, using \cref{eq:Khat_upper} together with Cauchy-Schwarz implies,
\begin{equation}\label{eq:cauchy_schwarz}
\langle \vx, \vx' \rangle \leq \norm{\vx}_\mZ \norm{\vx'}_{\mZ^{-1}} \stackrel{\mathrm{w.h.p}}{\leq} \frac{1}{\sqrt{mn}}\norm{\vx}_2 \norm{\vx'}_{\mZ^{-1}} .
\end{equation}
Applying the inequality above, we may write
\begin{equation} \label{eq:gtheta1}
\begin{split}
    \langle \bar\vg(G), \vtheta^{(J)} - \vtheta^0 \rangle & = \langle \bar\vg(G), \vtheta^{(J)} - \tilde\vtheta^{(J)} \rangle + \langle \bar\vg(G), \tilde\vtheta^{(J)} - \vtheta^0 \rangle\\
    & \stackrel{\mathrm{w.h.p}}{\leq} \frac{1}{\sqrt{m\eta}}\norm{\bar\vg(G)}_{\mZ^{-1}} \norm{\vtheta^{(J)} - \tilde\vtheta^{(J)}}_2 + \langle \bar\vg(G), \tilde\vtheta^{(J)} - \vtheta^0 \rangle\\
    & \stackrel{\mathrm{w.h.p}}{\leq} 2\norm{\frac{\bar\vg(G)}{\sqrt{m}}}_{\mZ^{-1}}  \sqrt{\frac{B}{m\eta\lambda}} + \langle \bar \vg(G), \tilde\vtheta^{(J)} - \vtheta^0 \rangle
\end{split}    
\end{equation} 

For the last inequality of \cref{eq:gtheta1} we have used Lemma \ref{lem:proxyGD}. Decomposing the second term of the right hand side in \cref{eq:gtheta1} gives,
\begin{equation} \label{eq:gtheta2}
    \begin{split}
         \langle \bar \vg(G), \tilde\vtheta^{(J)} - \vtheta^0 \rangle &= \langle \bar \vg(G), \frac{\mZ\vb}{\sqrt{m}}\rangle + \langle \bar\vg(G), \tilde\vtheta^{(J)} - \vtheta^0 - \frac{\mZ\vb}{\sqrt{m}} \rangle  \\
         & \stackrel{\mathrm{w.h.p}}{\leq} \frac{\bar\vg^T(G)\mZ\vb}{\sqrt{m}} + \frac{1}{\sqrt{\eta}}\norm{\frac{\bar\vg(G)}{\sqrt{m}}}_{\mZ^{-1}}\norm{\tilde\vtheta^{(J)} - \vtheta^0 - \frac{\mZ\vb}{\sqrt{m}}}_2 \\
         & \stackrel{\mathrm{w.h.p}}{\leq} \frac{\bar\vg^T(G)\mZ\vb}{\sqrt{m}} + \norm{\frac{\bar\vg(G)}{\sqrt{m}}}_{\mZ^{-1}}\sqrt{\frac{2B}{m\eta \lambda}} (1-\eta m \lambda)^{J/2} 
    \end{split}
\end{equation}
where the first inequlity is a consequence of \cref{eq:cauchy_schwarz}. The second inequality follows from the convergence of GD on the proxy loss $\tilde\cL$, given in Lemma \ref{lem:proxyGD}. By the definition of posterior mean and variance (Eq. \ref{eq:GPposteriors}) when the regularization parameter is set to $\lambda \leftarrow t\lambda$ we have,
\begin{align*}
    &\hat\mu_{t}(G) = \frac{\bar \vg^T(G)\mZ\vb}{\sqrt{m}},\\
    &\hat\sigma_{t}(G) = \norm{\frac{\bar\vg(G)}{\sqrt{m}}}_{\mZ^{-1}}.
\end{align*}
The final upper bound on $\gnn(G; \vtheta^{(J)}) - \hat\mu_t(G)$ follows from plugging in Equation \ref{eq:gtheta2} into Equation \ref{eq:gtheta1}, and applying Lemma \ref{lem:gnn_taylor}. Similarly, for the lower bound we have,
\begin{align} \label{eq:lowerbound1}
    - \gnn(G; \vtheta^{(J)}) &\stackrel{\mathrm{w.h.p}}{\leq} \langle \bar\vg(G),  \vtheta^0 - \vtheta^{(J)} \rangle + \tilde C L^3\left(\frac{B}{m \lambda}\right)^{2/3} \sqrt{m \log m}\\
    \label{eq:lowerbound2}
    \langle \bar\vg(G),  \vtheta^0 - \tilde\vtheta^{(J)} \rangle &\stackrel{\mathrm{w.h.p}}{\leq} -\hat\mu_t(G)  + \hat\sigma_t(G)\sqrt{\frac{2B}{m\eta \lambda}} (1-\eta m \lambda)^{J/2} \\
    \label{eq:lowerbound3}
     \langle \bar\vg(G),  \vtheta^0 - \vtheta^{(J)} \rangle &\stackrel{\mathrm{w.h.p}}{\leq} 2\hat\sigma_t(G) \sqrt{\frac{B}{m\eta\lambda}} +  \langle \bar\vg(G), \tilde\vtheta^{(J)} -  \vtheta^0\rangle
\end{align}
Where inequality \ref{eq:lowerbound1} holds by Lemma \ref{lem:GD_norm_bounds}, and the next two inequalities are driven similarly to equations \ref{eq:gtheta1} and \ref{eq:gtheta2}. The lower bound results by putting together equations \ref{eq:lowerbound1}-\ref{eq:lowerbound3}, and this concludes the proof. Note that we are implicitly taking a union bound over the 6 inequalities that all hold with high probability. The conditions of the used lemmas require that $m$ is picked at a $\poly(\log(N/\delta))$ rate, and a constant number of union bounds do not affect this rate.
\end{proof}
The next lemma gives a confidence interval over members of $\cH_{\gntk}$.

 \begin{lemma}[RKHS Confidence Interval from \citet{vakili2021optimal}]\label{thm:CI_kernelized}
 Let $f^* \in \cH_k$ with $\norm{f^*}_k \leq B$, and the observation noise to be sub-gaussian with parameter $\sigma^2$. Assume $H_t= \{ (\vx_\tau, y_\tau)\}$ with $(\vx_\tau)_{\tau \leq t}$ and $(\varepsilon_\tau)_{\tau \leq t}$ statistically independent. Then for a fixed input $\vx$, with probability greater than $1-\delta$, 
 \[
 \abs{f^*(\vx) - \mu_t(\vx)} \leq (B+ \frac{\sigma}{\sqrt{\lambda}}\sqrt{2\log(2/\delta)})\sigma_t(\vx).
 \]
 \end{lemma}
 
The following lemma shows that members of $\cH_{\gntk}$ are well described by the first-order Taylor approximation of a GNN around initialization. 

\begin{lemma}[\textbf{Approximation by a linearized GNN}] \label{lem:lin_gnn}
Let $f$ be a member of $\cH_{\gntk}$ with bounded RKHS norm $\norm{f}_{\gntk} \leq B$. Set $\delta \in (0,1)$ and let $N$ denote an upper bound on the possible number of nodes for a graph.  If 
$m = \cO\left(L^6\vert\cG\vert^4/\lambda_0^4\log(\vert \cG \vert^2 LN/\delta)\right)$,
then with probability greater than $1-\delta$, there exists $\vtheta^* \in \sR^p$ such that for all $G \in \cG$
\[
f(G) = \langle \ggnn(G; \vtheta^0), \vtheta^*\rangle, \quad \sqrt{m}\norm{\vtheta^*}_2 \leq \sqrt{2}B.
\]
\end{lemma}
\begin{proof}[\textbf{Proof of Lemma  \ref{lem:lin_gnn}}]
The proof follows the technique for Lemma 5.1 \citet{zhou2020neural} with some modifications.

From \cref{eq:lem_grad_gnn} proof of Lemma \ref{lem:gram_conv}, for $m = \cO(L^6/\epsilon^4\log(LN/\delta))$, and for $G_i$ and $G_j$ in the domain,
\[
        \abs{\gntk(G_i, G_j)- \ggnn^T(G_i; \vtheta^0)\ggnn(G_j; \vtheta^0)}  \leq (L+1)\epsilon
\]
with probability greater than $1-\delta$. Let $\mK_{\mathrm{full}}$ be the GNTK matrix calculated for all $G \in \cG$ and $\mgnn_{\mathrm{full}} = [\ggnn^T(G; \vtheta^0)]_{G \in \cG}$. Then applying a union bound over all $G \in \cG$, and setting $\delta \leftarrow \delta/\vert \cG\vert^2 $, if $m = \cO(L^6/\epsilon^4\log(\vert G \vert^2 LN/\delta))$ , then
\[
\norm{\mK_{\mathrm{full}} - \mgnn_{\mathrm{full}}^T\mgnn_{\mathrm{full}}/m}_F \leq \vert \cG\vert \epsilon
\]
with probability greater than $1-\delta$. Now applying this inequality when $\epsilon= \lambda_0/2\vert \cG\vert$, we get that if 
\[m = \cO(L^6\vert\cG\vert^4/\lambda_0^4\log(\vert G \vert^2 LN/\delta))\]
then $\norm{\mK_{\mathrm{full}}-\mgnn_{\mathrm{full}}^T\mgnn_{\mathrm{full}}/m}_F\leq \lambda_0/2$, with probability greater than $1-\delta$. Via the Triangle inequality we get that
\begin{equation}\label{eq:Gram_geq_GNTK}
\mgnn_{\mathrm{full}}^T\mgnn_{\mathrm{full}}/m \succcurlyeq \mK_{\mathrm{full}} - \norm{\mK_{\mathrm{full}}-\mgnn_{\mathrm{full}}^T\mgnn_{\mathrm{full}}/m}_F \succcurlyeq \mK_{\mathrm{full}} -\frac{\lambda_0}{2} \succcurlyeq \mK_{\mathrm{full}}/2 \succ 0.
\end{equation}
Since $\lambda_0 > 0$, $\bar\mG_{\mathrm{full}}$ is positive definite and may be decomposed as $\bar\mG_{\mathrm{full}} = \mP \mA \mQ^T$, where $\mP \in \sR^{p \times |\cG|}$, $\mP \in \sR^{|\cG| \times |\cG|}$ are unitary and $\mA \succ 0$. Let $\vf = [f(G)]_{G \in \cG}$ be the vector of function values.
We show that $\vtheta^* = \mP\mA^{-1}\mQ^T \vf$ satisfies the statement of the lemma. By definition of $\vtheta^*$,
\[
  \bar\mG_{\mathrm{full}}^T\vtheta^* = \mQ\mA\mP^T\mP\mA^{-1}\mQ^T\vf = \vf
\]
which implies for all $G \in \cG$, $\langle \ggnn(G; \vtheta^0),\vtheta^*\rangle = \gnn(G)$. As for the norm of $\vtheta^*$ we may write, 
\[
\norm{\vtheta^*}_2^2 \leq \vf^T\mQ\mA^{-2}\mQ^T = \vf^T (\mgnn_{\mathrm{full}}^T\mgnn_{\mathrm{full}}^T)^{-1} \vf \leq \frac{2}{m} \vf^T \mK_{\mathrm{full}}^{-1} \vf \leq \frac{2 B^2}{m}
\]
where the next to last inequality holds due to \cref{eq:Gram_geq_GNTK}, and the last inequality follows from $\norm{f}_{\gntk}^2 \leq B^2$.
\end{proof}
We are now ready to present the proof of our main confidence interval bound.
 
\begin{proof}[\textbf{Proof of \cref{thm:CI_gnn}}]
Consider \cref{thm:CI_kernelized}, when the kernel function is $\tilde k(G, G') = \bar\vg^T(G)\bar\vg(G')/m$ and choose the regularization parameter $t\lambda$.
Subsequently, the posterior mean and variance after observing $t$ samples will be $\hat \mu_t(G)$ and $\hat \sigma_t(G)$. Then this lemma  states that for $f \in \cH_{\tilde k}$ with a norm bounded by $B$, with probability greater than $1-\delta/2$,
\[
\vert f(G)-\tilde\mu_t(G) \vert \leq \hat \sigma_t \left( B + \frac{\sigma}{\sqrt{\lambda}}\sqrt{2\log 4/\delta} \right).
\]
By Lemma \ref{lem:lin_gnn}, for $m$ large enough the reward function can be written as
\[
f^*(G) = \langle \ggnn(G; \vtheta^0), \vtheta^*\rangle, \quad \sqrt{m}\norm{\vtheta^*}_2 \leq \sqrt{2}B.
\]
for all $G \in \cG$, indicating that $f^* \in \cH_{\tgntk}$ with $\norm{f}_{\tgntk}\leq \sqrt{2}B$. 
 Therefore, following \cref{thm:CI_kernelized} with probability greater than $1-\delta/2$, 
\begin{equation*}
       \vert f^*(G) - \hat\mu_t(G) \vert  \leq \hat \sigma_t \left( \sqrt{2}B + \frac{\sigma}{\sqrt{\lambda}}\sqrt{2\log 4/\delta} \right)
\end{equation*}
for some fixed $G$. 
Further, \cref{lem:CI_gnn_approxpostmean} bounds the difference between $\hat \mu_{t-1}(\cdot)$ and $\hat \mu_{t-1}(\cdot)$ with probability higher than $1-\delta/2$. Plugging in $\hat \mu_{t-1}$ and $\hat \sigma_{t-1}$ gives,
\begin{equation*}
    \begin{split}
        \vert \hat\mu_{t-1}(G)- f^*(G) \vert \leq \hat\sigma_{t}(G) \Bigg[ \sqrt{2}B + &\frac{\sigma}{\lambda}\sqrt{2\log4/\delta} +  \sqrt{\frac{2B}{m\eta\lambda}}\big( \sqrt{2} + (1-m\eta\lambda )^{J/2}\big) \Bigg] \\
        & + \tilde C L^3\left(\frac{B}{m \lambda}\right)^{2/3} \sqrt{m \log m}
    \end{split}
\end{equation*}
with probability greater than $1-\delta$. Setting $\delta \leftarrow \delta/\vert \cG\vert$ and taking a union bound over all $G \in \cG$ concludes the proof of \cref{thm:CI_gnn_formal}.
The informal version of the theorem is achieved by considering that $m = \poly(t)$ and omitting all the terms that are $o(t^{-1})$ with $t$.
\end{proof}
\subsection{GNN Helper Lemmas}
\begin{lemma}[\textbf{Norm concentration of Gram matrix and GNTK matrix at initialization}]\label{lem:gram_conv}
Set $\epsilon >0$, $\delta \in (0,1)$, and let $N = |V(G)|$ denote the number of nodes for every graph $G \in \cG$. For width $m = \Omega\big(L^6/\epsilon^4 \log(LNT^2/\delta)\big)$ in \cref{eq:def_gnn}, the following holds with probability at least $1-\delta$:
\begin{equation*}
    \norm{\mgntk- \mgnn^T\mgnn/m}_F \leq  T \epsilon.
\end{equation*}
\end{lemma}

\begin{proof}[\textbf{Proof of Lemma \ref{lem:gram_conv}}]
We make use of the connection between the GNN as defined in \cref{eq:def_gnn} and a fully-connected neural network. In particular, let $\nn(\vx; \vtheta): \mathbb{R}^d \rightarrow \mathbb{R}$ be a fully-connected network, with $L$ hidden layers of equal width $m$, and ReLU activations as defined in \cref{eq:def_nn}. Then, we have
\begin{equation*} 
    \gnn(G; \vtheta) = \frac{1}{\sqrt{N}}\sum_{v \in V(G)} \nn(\bar \vh_{v}; \vtheta).
\end{equation*}
Let $\mgnn$ 
be a matrix with $T$ columns where each column $i \in [T]$ contains gradient vector $\ggnn(G_i; \vtheta^0)$. Then, the matrix $\mgnn^T \mgnn$ is a $T \times T$ matrix and represents the Gram matrix of the network for the parameters $\vtheta^0$. For all $i, j \leq T$, we have
\begin{equation}\label{eq:mgnn_sum_nn}
    [\mgnn^T\mgnn]_{i,j} = \frac{1}{N} \sum_{v=1}^N \vg^T(\bar \vh_v)\vg(\bar \vh_{v'}),
\end{equation}
where $\vg(\vx)=\nabla_\vtheta \nn(\vx; \vtheta^0)$ denotes the gradient of $\nn$ at initialization.
It follows by the definition of the neural tangent kernel  that
\begin{equation}\label{eq:gntk_sum_nn}
\gntk(G_i, G_j) = \frac{1}{N}\sum_{v=1}^N \ntk(\bar \vh_v, \bar \vh_{v'}). 
\end{equation}
For some fixed $\epsilon>0$ and $\delta \in (0,1)$, the result of  \citet[Theorem 3.1]{arora2019exact} states that when $m = \Omega(L^6 /\epsilon^4 \log (L/\delta))$, the following holds for any $\vx, \vx'$ with unit norms and probability at least $1-\delta$:
\begin{equation}\label{eq:arora}
    \abs{\ntk(\vx, \vx')-\vg^T(\vx)\vg(\vx')/m} \leq (L+1)\epsilon.
\end{equation}
\looseness-1 Next, by using equations \ref{eq:gntk_sum_nn} and \ref{eq:mgnn_sum_nn}, and the triangle inequality for any two input graphs $G_i$, $G_j$ we get
\begin{equation}\label{eq:lem_grad_gnn}
        \abs{\gntk(G_i, G_j)- \ggnn^T(G_i; \vtheta^0)\ggnn(G_j; \vtheta^0)}  \leq \frac{1}{N} \sum_{v=1}^N \abs{ \ntk(\bar \vh_v, \bar \vh_{v'}) -  \vg^T(\bar \vh_v)\vg(\bar \vh_{v'})}.
\end{equation}
Then, since $\bar \vh_v \in \mathbb{S}^{d-1}$ for every node $v$ irrespective of the corresponding graph, we can use \cref{eq:arora} together with the union bound over all $(\bar \vh_{G,u}, \bar \vh_{G', u})$ pairs and $m = \Omega(L^6 /\epsilon^4 \log (LN/\delta))$, to obtain that for any $G_i$ and $G_j$ with probability at least $1-\delta$:
\begin{equation*}
            \abs{\gntk(G_i, G_j)-\ggnn^T(G_i; \vtheta^0)\ggnn(G_j; \vtheta^0)/m} \leq (L+1) \epsilon.
\end{equation*}
To arrive at the main result we consider the difference in the Frobenius norm:
\begin{equation*}
    \norm{\mgntk- \mgnn^T\mgnn/m}_F = \sqrt{\sum_{i,j \leq T} \Big( \gntk(G_i, G_j)-\ggnn^T(G_i; \vtheta^0)\ggnn(G_j; \vtheta^0)/m\Big)^2}.
\end{equation*}
By setting $\epsilon \leftarrow \epsilon/(L+1)$ and again applying the union bound over each $(G_i, G_j)$ pair, for $m = 
\Omega(L^6/\epsilon^4 \log(LNT^2/\delta))$, the following holds with probability at least $1-\delta$:
\begin{equation*}
    \norm{\mgntk- \mgnn^T\mgnn/m}_F \leq  T \epsilon.
\end{equation*}
\end{proof}

\begin{lemma}[\textbf{Gradient descent norm bounds}]\label{lem:GD_norm_bounds}
Consider the fixed set $\{G_i\}_{i\leq t}$ of inputs. Let $\mgnn =[ \ggnn^T(G_i; \vtheta^0)]^T_{i \leq t}$ be the matrix of gradients at initialization and $\mgnn^{(j)} =[ \ggnn^T(G_i; \vtheta^{(j)})]^T_{i \leq t}$. The vector of network outputs after the $j$-th update is denoted by $\vgnn^{(j)} = [\gnn(G_i; \vtheta^{(j)})]_{i \leq t}$. Assume $\tau$ is set such that $\vert \vert \vtheta^{(j)} - \vtheta^0\vert\vert_2 \leq \tau$ for all $j \leq J$ . If $m = \poly \Big(t, L, \lambda^{-1},  \log(N/\delta)\Big)$, then with probability greater than $1-\delta$,
\begin{align}
   \norm{\mgnn}_F & \leq C_1 \sqrt{tmL} \label{eq:bound_Fnorm_G} \\
   \norm{\mgnn-\mgnn^{(j)}}_F & \leq C_2  \tau^{1/3}L^{7/2} \sqrt{tm\log m} \label{eq:bound_G_normdif}\\
   \norm{\vgnn^{(j)}-\mgnn^{(j)}(\vtheta^{(j)}-\vtheta^0)}_2 & \leq C_3\tau^{4/3}L^3
\sqrt{tm\log m} \label{eq:bound_fdiff}
\end{align}
for some constants $C_1, \,C_2,\,C_3$.
\end{lemma}
\begin{proof}[\textbf{Proof of Lemma \ref{lem:GD_norm_bounds}}] We follow the recipe introduced in \citet{zhou2020neural}, and reproduce gradient norm bounds for when the network is a GNN.

From Lemma B.3 \citet{cao2019generalization}, we get $\norm{\ggnn(G_i; \vtheta^0)}_2 \leq \bar C \sqrt{mL}$ with probability of at least $1-\delta/3$. By the definition of Frobenius norm, it follows,
\begin{equation*}
    \norm{\mgnn}_F \leq \sqrt{t} \max_{i \leq t} C \norm{\ggnn(G_i; \vtheta^0)}_2 \stackrel{\mathrm{w.h.p}}{\leq} C \sqrt{tmL}.
\end{equation*}
For \cref{eq:bound_G_normdif} we may write,
\begin{equation*}
  \begin{split}
      \norm{\mgnn^{(j)}-\mgnn}_F & \leq \sqrt{t} \max_{i \leq t} \norm{\ggnn(G_i; \vtheta^{(j)})-\ggnn(G_i; \vtheta^0)}_2 \\
      & \leq\frac{\sqrt{t}}{N} \max_{i \leq t} \sum_{u \in V(G_i)}  \norm{\vg(\bar \vh_v; \vtheta^{(j)})-\vg(\bar \vh_v; \vtheta^0)}_2\\
      & \stackrel{\mathrm{w.h.p}}{\leq} \frac{\sqrt{t}}{N} \max_{i \leq t}  \sum_{u \in V(G_i)}\tilde C_2 \tau^{1/3}L^3\sqrt{ \log m}  \norm{\vg(\bar \vh_v; \vtheta^0)}_2\\
      & \stackrel{\mathrm{w.h.p}}{\leq} C_2 \tau^{1/3} L^{7/2}\sqrt{tm\log m}
  \end{split}  
\end{equation*}
with probability greater than $1-\delta/3$, where the next to last inequality holds by Lemma B.5 \citet{zhou2020neural} and the last inequality follows directly from Lemma B.6 \citet{zhou2020neural}. 

As for \cref{eq:bound_fdiff}, by definition of the single-\textsc{BLOCK} GNN and Lemma B.4 \citet{zhou2020neural},
\begin{equation*}
    \begin{split}
        \norm{\vgnn^{(j)}-\mgnn^{(j)}(\vtheta^{(j)}-\vtheta^0)}_2 &\leq 
        \sqrt{t}  \max_{i \leq t} \abs{\gnn(G_i; \vtheta^{(j)})-\langle \ggnn(G_i; \vtheta^{(j)}), \vtheta^{(j)}-\vtheta^0\rangle}\\
        & \leq \frac{\sqrt{t}}{N} \max_{i \leq t}  \sum_{u \in V(G_i)} \abs{f(\bar \vh_v; \vtheta^{(j)})-\langle \vg(\bar \vh_v; \vtheta^{(j)}), \vtheta^{(j)}-\vtheta^0\rangle}\\
        & \stackrel{\mathrm{w.h.p}}{\leq} C_3 \tau^{4/3}L^3 \sqrt{tm\log m}
    \end{split}
\end{equation*}
with probability greater than $1-\delta/3$.
\end{proof}

\begin{lemma}[\textbf{Convergence properties of the proxy optimization problem}]\label{lem:proxyGD}
Let the sequence $(\tilde \vtheta^{(j)})$ denote the gradient descent updates on the following loss function,
\[
\tilde\cL(\vtheta) =  \frac{1}{t}\sum_{i\leq t} \big(\langle\ggnn(G_i; \vtheta^0),\vtheta-\vtheta^0\rangle - y_i\big)_2^2     + m \lambda \norm{\vtheta-\vtheta^0}_2^2.
\]
then if $m =\poly(T, L,  B, \vert \cG\vert,\lambda_0^{-1},\lambda^{-1}, \log(N/\delta))$ and the learning rate $\eta \leq C(mL+m\lambda)^{-1}$
\begin{align*}
    &\norm{\tilde \vtheta^{(j)} - \vtheta^0}_2 \leq \sqrt{B/m\lambda},\\
    & \norm{\tilde\vtheta^{(j)} - \vtheta^0 - \left(\lambda\mI + \mgnn^T\mgnn/(tm) \right)^{-1} \mgnn^T\vy/(tm)}_2 \leq (1-\eta m \lambda)^{j/2} \sqrt{2B/m \lambda}.\\
\end{align*}
\end{lemma}
\begin{proof}[\textbf{Proof of Lemma \ref{lem:proxyGD}}]
This lemma adapts Lemma D.8 \citet{kassraie2021neural} to our setting. We repeat the proof for the sake of completeness. Note that $\tilde \cL$ is $m\lambda$-strongly convex, and $C(mL+m\lambda)$-smooth, 
since
\[
\nabla^2\tilde \cL =\frac{ 2\mgnn^T\mgnn}{t} + 2m\lambda\mI \leq 2(\frac{\norm{\mgnn}_F^2 }{t}+ m\lambda)\mI \leq C (mL+m\lambda) \mI,
\]
where the second inequality follows from Lemma \ref{lem:GD_norm_bounds}. Strong Convexity of $\tilde\cL$ guarantees a monotonic decrease of the loss if the learning rate is smaller than the smoothness coefficient inversed. Therefore, 
\begin{align*}
    m\lambda \norm{\tilde\vtheta^{(j)} - \tilde\vtheta^0}_2^2 & \leq m\lambda \norm{\tilde\vtheta^{(j)} - \tilde\vtheta^0}_2^2 + \frac{1}{t}\norm{\mgnn(\tilde\vtheta^{(j)} - \tilde\vtheta^0)-\vy}_2^2\\
    & \leq m\lambda\norm{\tilde\vtheta^0 - \tilde\vtheta^0}_2^2 + \frac{1}{t}\norm{\mgnn(\tilde\vtheta^0 - \tilde\vtheta^0)-\vy}_2^2\\
    & \leq \frac{\norm{\vy}^2_2}{t}\\
    & \leq B
\end{align*}
From the RKHS assumption, the true reward is bounded by $B$ and hence the last inequality follows since the size of the training set is $t$.

Gradient descent on smooth and strongly convex functions converges to optima if the learning rate is smaller than the smoothness coefficient inversed. Under this condition the minima of $\tilde\cL$ is unique and has the closed form
\[
\tilde\vtheta^* = \vtheta^0 +  \left( \lambda \mI + \mgnn^T\mgnn/(mt) \right)^{-1} \mgnn^T\vy/(mt)
\]
\looseness-1 Having set $\eta \leq C(mL+m\lambda)^{-1}$, we get that $\tilde\vtheta^{(j)}$ converges to $\tilde\vtheta^*$ with the following exponential rate,
\begin{align*}
    \norm{\tilde\vtheta^{(j)} - \vtheta^0 - \left(\lambda\mI + \mgnn^T\mgnn/(mt) \right) ^{-1} \mgnn^T\vy/(mt)}_2^2 & \leq (1-\eta m \lambda)^{(j)} \frac{2}{m\lambda}\left(\tilde\cL(\vtheta^0)-\tilde\cL(\tilde\vtheta^*)\right)\\
    & \leq \frac{2(1-\eta m \lambda)^j}{m \lambda} \frac{\norm{y}_2^2}{t}\\
    & \leq \frac{2B(1-\eta m \lambda)^j}{m \lambda}.
\end{align*}
\end{proof}
\begin{lemma}[\textbf{Gradient descent parameters bound}]\label{lem:GD_theta_bound}
Let the sequence $\vtheta^{(J)}$ denote the $J$-th gradient descent update on the GNN loss,
\[
\tilde\cL(\vtheta) =  \frac{1}{t}\sum_{i\leq t} \big(\gnn(G_i; \vtheta^0) - y_i\big)_2^2     + m \lambda \norm{\vtheta-\vtheta^0}_2^2.
\]
If $m =\poly(T, L,  B, \vert \cG\vert,\lambda_0^{-1},\lambda^{-1}, \log(N/\delta))$ and $\eta \leq C(mL+m\lambda)^{-1}$ for some $C$, then with probability greater than $1-\delta$,
\[
\norm{\vtheta^{(J)}-\vtheta^0}_2 \leq 2\sqrt{B/m \lambda}.
\]
\end{lemma}
\begin{proof}[\textbf{Proof of Lemma \ref{lem:GD_theta_bound}}]
Following \citet{zhou2020neural} we introduce the sequence $(\tilde \vtheta^{(j)})$ which denotes the gradient descent updates on the following proxy loss,
\[
\tilde\cL(\vtheta) = \frac{1}{t} \sum_{i\leq t} \big(\langle\ggnn(G_i; \vtheta^0),\vtheta-\vtheta^0\rangle - y_i\big)_2^2     + m \lambda \norm{\vtheta-\vtheta^0}_2^2.
\]
By Lemma \ref{lem:proxyGD},
\begin{align*}
     \norm{\tilde\vtheta^{(J)} - \vtheta^0}_2 \leq \sqrt{\frac{B}{m\lambda}}
\end{align*}
It remains to show that $\vert \vert \vtheta^{(J)} - \tilde \vtheta^{(J)}\vert \vert_2 \leq \sqrt{B/m \lambda }$, which concludes the proof due to triangle inequality. By writing out the gradient descent updates of the two sequences we get,
\begin{equation*}
    \begin{split}
        \norm{\vtheta^{(j+1)}-\tilde\vtheta^{(j+1)}}_2 & = \bigg\vert\bigg\vert(1-\eta m \lambda)(\vtheta^{(j)}-\tilde \vtheta^{(j)}) - \frac{\eta}{t}(\mgnn^{(j)}-\mgnn)^T(\vgnn^{(j)}-\vy) \\
        & \quad\quad- 
        \frac{\eta}{t}\mgnn^T \big( \vgnn^{(j)} - \mgnn(\vtheta^{(j)}-\vtheta^0)+\mgnn(\vtheta^{(j)}-\tilde\vtheta^{(j)})\big)\bigg\vert\bigg\vert_2\\
        & \leq \frac{\eta}{t} \norm{(\mgnn^{(j)}-\mgnn)}_2\norm{(\vgnn^{(j)}-\vy)}_2 + \frac{\eta}{t}\norm{\mgnn}_2\norm{\vgnn^{(j)} - \mgnn(\vtheta^{(j)}-\vtheta^0)}_2 \\
        &\quad \quad+ \norm{\mI - \eta\big(m\lambda \mI + \mgnn\mgnn^T/t\big)}_2\norm{ \vtheta^{(j)} - \tilde\vtheta^{(j)}}_2
    \end{split}
\end{equation*}
We bound each term separately. In the rest of the proof, Lemma \ref{lem:GD_norm_bounds} is always used with $\tau = \sqrt{B/m\lambda}$. Lemma C.3 \citet{zhou2020neural} directly holds for $\gnn$, and states that $\vert \vert \vgnn^{(j)}-\vy\vert\vert_2 \leq (B+1)\sqrt{t}$. Then \cref{eq:bound_G_normdif} Lemma \ref{lem:GD_norm_bounds} gives,
\[
\frac{\eta}{t}\norm{(\mgnn^{(j)}-\mgnn)}_2\norm{(\vgnn^{(j)}-\vy)}_2 \stackrel{\mathrm{w.h.p}}{\leq} \eta C_1  \left( \frac{B}{m\lambda}\right)^{1/6} L^{7/2}(B+1)\sqrt{m\log m}.
\]
Recall that $\gnn(G; \vtheta^0) = 0$ by design. Then for the second term, using Equations \ref{eq:bound_Fnorm_G} and \ref{eq:bound_fdiff} from Lemma \ref{lem:GD_norm_bounds},
\[
\frac{\eta}{t}\norm{\mgnn}_2\norm{\vgnn^{(j)} - \mgnn(\vtheta^{(j)}-\vtheta^0)}_2 \stackrel{\mathrm{w.h.p}}{\leq} \eta C_2 \left( \frac{B}{m\lambda}\right)^{2/3}L^{7/2}m\sqrt{\log m}.
\]
As for the last term, first note that by \cref{eq:bound_Fnorm_G} Lemma \ref{lem:GD_norm_bounds}, with high probability
\[
\eta\big(m\lambda \mI + \mgnn\mgnn^T/t\big) \stackrel{\mathrm{w.h.p}}{\preccurlyeq} \eta (m\lambda \mI + C_1mL\mI) \preccurlyeq \mI
\]
where the last inequality holds since $\eta$ is chosen to be small enough. Therefore,
\[
\norm{\mI - \eta\big(m\lambda \mI + \mgnn\mgnn^T\big)}_2\norm{ \vtheta^{(j)} - \tilde \vtheta^{(j)}}_2 \stackrel{\mathrm{w.h.p}}{\leq} (1-\eta m\lambda)\norm{\vtheta^{(j)}-\tilde\vtheta^{(j)}}_2.
\]
We put the three terms back together and unroll the recursive inequality. Then if $m$ is picked to be large enough at the above stated polynomial rate,
\[
\norm{\vtheta^{(j)}-\tilde \vtheta^{(j)}}_2 \stackrel{\mathrm{w.h.p}}{\leq} \sqrt{\frac{B}{m\lambda}}.
\]
\end{proof}
The next lemma shows that the first order approximation of a GNN at initialization can still describe the network after it has been trained with gradient descent for $J$ steps.
\begin{lemma}[\textbf{Taylor approximation of a trained GNN}] \label{lem:gnn_taylor}
 If $m = \poly(T, L, B, \vert \cG\vert, \lambda_0^{-1}, \lambda^{-1},\log(N/\delta))$ and for some constant $C$, $\eta = C(mL+m\lambda)^{-1}$,  then \looseness-1
\[
\abs{\gnn(G_t; \vtheta^{(J)}) - \gnn(G_t; \vtheta^0) - \langle \ggnn(G_t; \vtheta^0), \vtheta^{(J)} - \vtheta^0 \rangle} \leq \tilde C L^3\left(\frac{B}{m \lambda}\right)^{2/3} \sqrt{m \log m}
\]
with some constant $\tilde C$ and any $t \leq T$, with probability greater than $1-\delta$.
\end{lemma}
\begin{proof}[\textbf{Proof of lemma \ref{lem:gnn_taylor}}]
By Lemma 4.1 \citet{cao2019generalization}, if $m = \poly(T, L,  B, \vert \cG\vert,\lambda_0^{-1},\lambda^{-1}, \log(1/\delta))$ and $\eta$ is set according to the statement of the lemma, then for a fixed $\vx$ with probability greater than $1-\delta$,
\[
\abs{\nn(\vx; \vtheta^{(J)}) - \nn(\vx; \vtheta^0)- \langle g(\vx; \vtheta^0), \vtheta^{(J)} - \vtheta^0\rangle} \leq C \tau^{4/3}L^3 \sqrt{m \log m}
\]
where $\norm{\vtheta^{(J)} - \vtheta^0}_2 \leq \tau$. We use this inequality with $\vx = (\bar \vh_v^{(i)})$ for all $u \in V(G_i)$. Setting $\delta \leftarrow \delta/2N$ and applying the union bound gives 
\begin{align*}
      \Bigg\vert \sum_{u \in V(G_i)} \nn(\bar \vh_v^{(i)}; \vtheta^{(J)}) - \sum_{u \in V(G_i)} \nn(\bar \vh_v^{(i)}; \vtheta^0) - & \langle \sum_{u \in V(G_i)} \vg(\bar \vh_v^{(i)}; \vtheta^0),\vtheta^{(J)} - \vtheta^0\rangle \Bigg\vert \\
      & \leq C N \tau^{4/3}L^3 \sqrt{m \log m} 
\end{align*}
Therefore, if $m = \poly(T, L, B, \vert \cG\vert,\lambda_0^{-1},\lambda^{-1}, \log(N/\delta)))$ with probability greater than $1-\delta$,
\begin{equation} \label{eq:taylor_gnn_eta}
         \abs{f(G_i; \vtheta^{(J)}) - f(G_i; \vtheta^0) - \langle \ggnn(G_i; \vtheta^0),\vtheta^{(J)} - \vtheta^0\rangle} \leq C \tau^{4/3}L^3 \sqrt{m \log m}   
\end{equation}
It remains to bound $\vert\vert \vtheta^{(J)} - \vtheta^0\vert\vert_2$ in order to calculate $\eta$ in Equation \ref{eq:taylor_gnn_eta}. From Lemma \ref{lem:GD_theta_bound}, we have $\vert\vert \vtheta^{(J)} - \vtheta^0\vert\vert_2 \leq 2\sqrt{B/m\lambda}$. Setting $\delta \leftarrow \delta/T$  and taking a union bound over all $t \leq T$ concludes the proof. Note that the added $\log T$ term from union bound does not change rate for $m$, since it is already growing polynomially with $T$. 
\end{proof}

\section{Experiments}
We include the details of the experiments in \cref{sec:experiments}, together with the supplementary plots. 
\subsection{Synthetic Permutation Invariant Datasets}
To test our permutation invariant additive model, we pick the GNTK as the kernel function and create $18$ datasets that inherit this structure. 
As explained in \cref{sec:experiments}, each dataset consists of a finite domain of size $\cG = 10000$ together with a reward function, both of which are generated randomly. 
The domains are sets of Erd\H{o}s-R\'{e}nyi random graphs, where each graph has $N$ nodes, and between each two nodes there exists an edge with probability $p$. The node features are i.i.d. $d$-dimensional standard Gaussian vectors. 

For every domain, we sample a random reward function. We use $\mathrm{GP}(0, \gntk)$ as a prior, and sample $f$ from its posterior GP. The posterior is calculated using a small random dataset $(G_i, y_i)_{i \leq 5}$, where $y_i$ are drawn independently from $\cN(0,1)$ and $G_i$ are randomly chosen from $\cG_{p,N}$. 
We choose the posterior GP over the prior as it produces somewhat smoother samples. 
We note that functions drawn from this GP {\em do not} reside in $\Hgntk$.
\cref{tab:dataset} shows the characteristics of the datasets, which will be released together with the code to generate them from scratch.

\begin{table}[ht]
    \centering
    \begin{tabular}{c|c c  c}
         & $N=5$ & $N=20$ & $N=100$\\
         \hline
        $p=0.05$ & $d \in \{ 10,100\}$ & $d \in \{ 10,100\}$ & $d \in \{ 10\}$\\
        $p = 0.2$ & $d \in \{ 10,100\}$ & $d \in \{ 10,100\}$ & $d \in \{ 10,100\}$\\
        $p = 0.95$ & $d \in \{ 10,100\}$ & $d \in \{ 10,100\}$ & $d \in \{ 10,100\}$ \\
    \end{tabular}
    \caption{Parameters of the synthetic datasets}
    \label{tab:dataset}
\end{table}

\subsection{Practical Details} \label{app:training}
The python code to our algorithms, bandit environment, and experiments will be released. 

\textbf{Algorithm. } There are some differences between how we utilize the algorithm in practice and the pseudo-code in\cref{alg:gnnalg}. We list these modifications for transparency. 
\begin{itemize}
    \item When calculating $\hat \sigma_{t-1}$ we approximate $\hat \mK_{t-1}$ with its diagonal so that the matrix inversion takes $o(t)$ operations.
    \item  \gnnus suggests to discard data from previous episodes, so that the decisions are non-adaptive. In practice we keep the history for training the network. 
    \item  We set all $T_e=1$. 
    \item  Only from  $t \geq T_2 = 80$ we follow \cref{eq:potential_maximizers_def} and intersect the sets of plausible maximizers. For the first $T_2$ steps construct them via
 \begin{equation*}
     \mathcal{G}_{e+1} \leftarrow \big\lbrace G \in \mathcal{G}: \gnn(G;\vtheta^{(J)}_{e}) + \beta_{T_e} \hat{\sigma}_{T_e}(G) \geq  \max_{G \in \mathcal{G}} \big( \gnn(G;\vtheta^{(J)}_{e}) - \beta_{T_e} \hat{\sigma}_{T_e}(G)\big)  \big \rbrace.
 \end{equation*}
\end{itemize}

\textbf{Network Architecture.} We set the width of all architectures to $m=2048$ and depth to $L=2$. This combination is picked primarily to keep computations light, while somewhat adhering to the theoretical setup. To calculate $\hat \sigma_{t-1}$ we approximate the gram matrix $\mG^T\mG$ with its diagonal, which gets worse as the number of network parameters grow. The picked values for $m$ and $L$ producing a descriptive network, and allow us to use this diagonal approximation with a negligible cumulative error.\looseness-1

\textbf{Graph Neural Tangent Kernel} To implement this kernel function, we use the NTK class from the Neural Tangents library \cite{neuraltangents2020}, and sum the base NTK via \cref{eq:gntk_vs_ntk}. 
This library offers the tangent kernels of every network architecture, however it is unclear how the kernel is derived for a GNN, therefore we use our own expression. 

\textbf{Initialization.} We initialize the networks by directly following the definition of \cref{eq:def_gnn}. The scaling with $1/\sqrt{m}$ is crucial in activating networks in the lazy regime. If this condition is not met, the confidence sets $[\gnn(\cdot; \vtheta)\pm \hat\sigma_{t-1}(\cdot)]$ may not be valid, since $\hat\sigma_{t-1}$ no longer accurately describes the posterior variance of $\gnn$.

\textbf{Training. } When analyzing the training dynamics of $\gnn$, we consider SGD on the $\ell_2$-regularized loss. In practice, however, we train the network with the Adam optimizer \citep{kingma2014adam} from \textsc{PyTorch} \citep{paszke2017automatic}, and without weight decay. The learning rate is set to $\eta = 0.001$. We allow $T_0=40$ steps of random exploration, to mimic some form of pre-training. The random exploration steps are included in our regret plots.
For the first $T_1 = 100$ steps, we train the network from scratch (using the same initialization $\vtheta_0$) at every step $t$, as described by the algorithm, and then in batches of $T_B= 20$ just to keep computations light.
At every step $t$ we run the Adam optimizer for $J_t$ gradient descent steps, where $J_t$ is calculated via the following stopping criteria
\begin{align*}\label{eq:GD_stopping}
   J_t = \min  \quad & J\\
    \text{s.t.}  \quad &\cL(\vtheta^{(J)}_{t-1}) \leq \cL_0\quad   \text{or,} \quad \frac{\Delta\cL^{(J-1)}_{t-1} - \Delta\cL^{(J)}_{t-1}}{\Delta \cL^{(J-1)}_{t-1}} \leq \delta_0
\end{align*}
where we set $\cL_0 =10^{-4}$, $\delta_0 = 10^{-3}$, and
\[
\Delta\cL^{(J)}_{t-1} := \cL(\vtheta^{(J)}_{t-1})-\cL(\vtheta^{(J-1)}_{t-1}).
\]
The above criterion targets both value of the loss function and the rate at which it is decaying. Effectively, this rule stops training if either the loss is lower than a threshold $\cL_0$, or if the loss has plateaued, i.e. the relative change in the the loss is lower than a threshold $\delta_0$. Roughly put, the two conditions on value and decay of the loss, cause the training algorithm to run longer for larger $t$ and prevent over-fitting when $t$ is small.
The hyperparameters of the optimizer, i.e., $\eta, \delta_0, \cL_0, T_B, T_0$, and $T_1$ are selected by hand and not automatically tuned.

\subsection{\gnnucb \& \nnucb} \label{app:ucb_algs}
In \cref{sec:experiments}, we compare \gnnus with \nnus, \gnnucb and \nnucb as baselines.
The pseudo-code is laid out in \cref{alg:gnnucb} and \cref{alg:nnucb}.

\begin{algorithm}[ht] 
    \caption{\label{alg:gnnucb} \gnnucb
    }
\DontPrintSemicolon
\KwInput{$m,\,J,\,\eta,\,\lambda,\,\beta_t,\, T$}
\init{ network parameters to a random $\vtheta^0$, and $\hat\mK_0 = \sigma^2 \mI$.}{}
\For{$t = 1 \cdots T$}{
 \For{$G \in \cG$}
{
$\hat{\sigma}^2_{t-1}(G) \leftarrow \ggnn^T(G; \vtheta^0) \hat{\bm{K}}^{-1}_{t-1} \ggnn(G; \vtheta^0)/m$\\
$U_{G,t} \leftarrow \gnn(G; \vtheta^{(J)}_{t-1}) + \beta_t\hat\sigma_{t-1}(G)$
}
$G_t = \argmax_{G\in \cG} U_{G,t}$\\
Select $G_t$ and append the rewards vector $\vy_t$ by the observed reward. \\
Set $\hat\mK_{t} \leftarrow \lambda \bm{I} + \sum_{i \leq t} \ggnn(G_i; \vtheta^0)\ggnn^T(G_i; \vtheta^0)/mt$ \\
Calculate $\vtheta_{t}^{(J)} = \text{TrainGNN}\left(m , J, \eta, \lambda, \vtheta^0, (G_i, y_i)_{i \leq t}\right)$\\
}
\end{algorithm}

\begin{algorithm}[ht] 
    \caption{\label{alg:nnucb} \nnucb
    }
\DontPrintSemicolon
\KwInput{$m,\,J,\,\eta,\,\lambda,\,\beta_t,\, T$}
\init{ network parameters to a random $\vtheta^0$, and $\hat\mK_0 = \sigma^2 \mI$.}{}
\For{$t = 1 \cdots T$}{
 \For{$G \in \cG$}
{
$\hat{\sigma}^2_{t-1}(G) \leftarrow \gradnn^T(\bar\vh_G; \vtheta^0) \hat{\bm{K}}^{-1}_{t-1} \gradnn(\bar\vh_G; \vtheta^0)/m$\\
$U_{G,t} \leftarrow \nn(\bar\vh_G; \vtheta^{(J)}_{t-1}) + \beta_t\hat\sigma_{t-1}(G)$
}
$G_t = \argmax_{G\in \cG} U_{G,t}$\\
Select $G_t$ and append the rewards vector $\vy_t$ by the observed reward. \\
Set $\hat\mK_{t} \leftarrow \lambda \bm{I} + \sum_{i \leq t} \gradnn(\bar\vh_{G_i}; \vtheta^0)\gradnn^T(\bar\vh_{G_i}; \vtheta^0)/mt$ \\
Calculate $\vtheta_{t}^{(J)} = \text{TrainNN}\left(m , J, \eta, \lambda, \vtheta^0, (\bar\vh_{G_i}, y_i)_{i \leq t}\right)$\\
}
\end{algorithm}

\begin{algorithm}[ht] 
\DontPrintSemicolon
\KwInput{$m,\,J,\,\eta,\,\lambda,\,\vtheta^0,\,(\bar\vh_{G_i}, y_i)_{i < t}$}
Define 
    $\cL(\vtheta) = \frac{1}{t}\sum_{i < t } \big(\nn(\bar\vh_{G_i}; \vtheta) - y_i\big)^2 + m \lambda \norm{\vtheta - \vtheta^0}_2^2$\\
Initialize $\vtheta^{(0)} = \vtheta^0$ \\
 \For{$j = 1, \dots, J$}
{
$\vtheta^{(j)} = \vtheta^{(j-1)} - \eta \nabla\cL(\vtheta^{(j-1)})$
}
\KwOutput{$\vtheta^{(J)}$}
\caption{TrainNN}
\end{algorithm}

\begin{figure}[ht] 
    \centering
    \includegraphics[width = \linewidth]{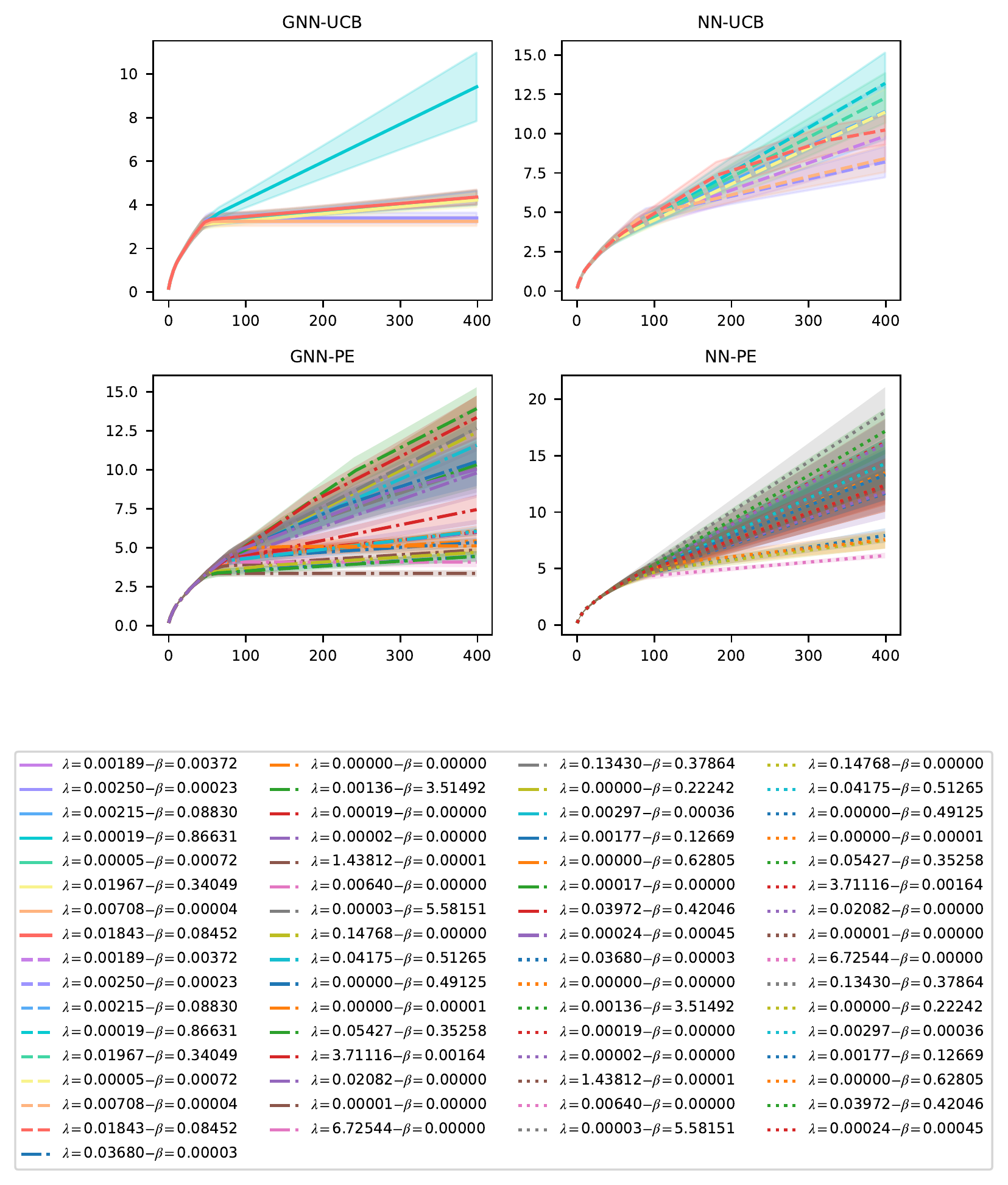}
    \caption{Results of hyper-parameter search for all algorithms. The GNN methods then to perform well for many configurations of $\lambda$ and $\beta$.}
    \label{fig:hyper_d10}
\end{figure}

\begin{figure}[ht] 
    \centering
    \includegraphics[width=\linewidth]{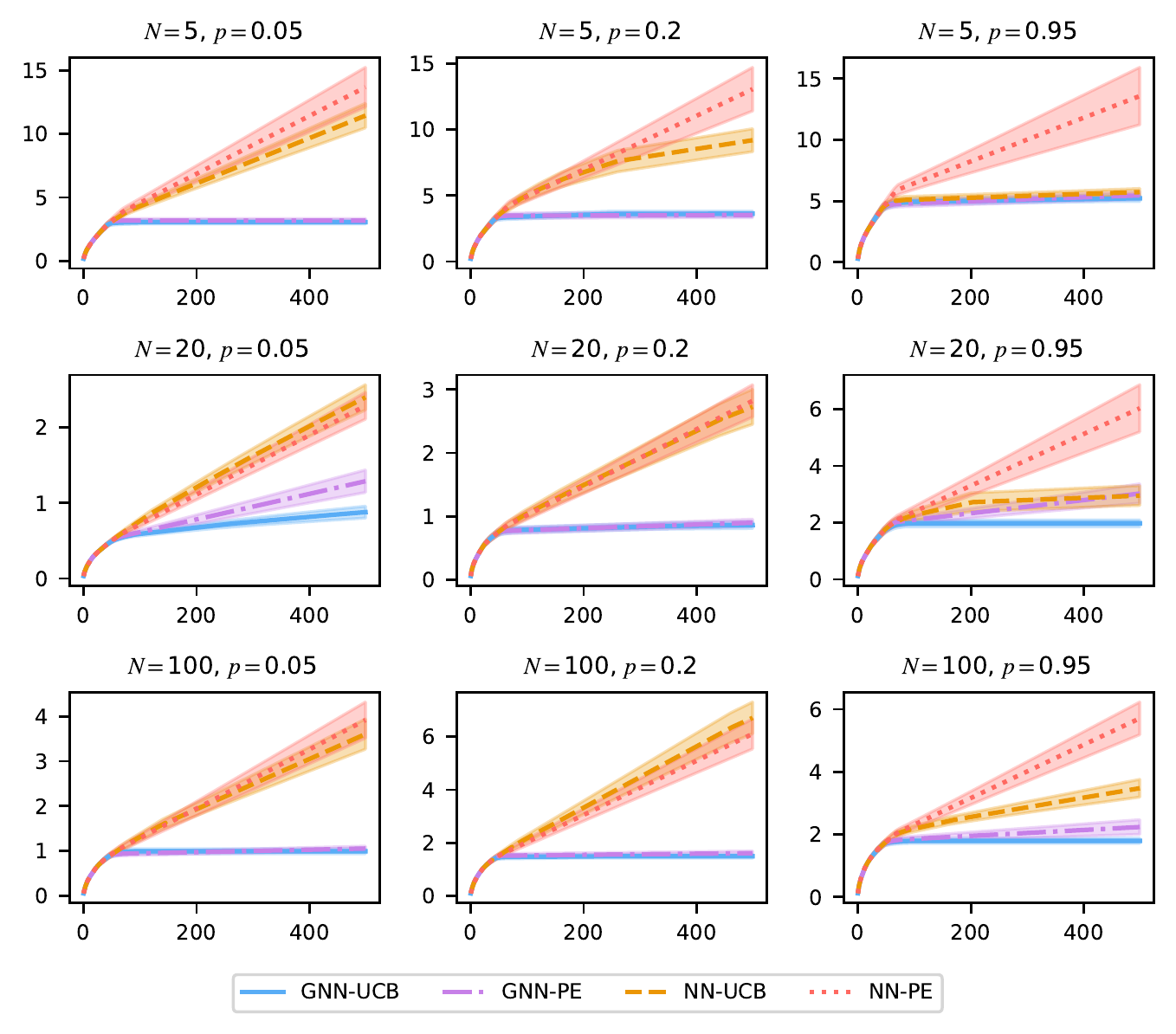}
    \caption{Comparing performance of \gnnus, \gnnucb, \nnus, and \nnucb for all dataset configurations.
    The GNN methods consistently outperform NN methods.
    Hyper-parameter tuning is done only for $N=5, p=0.05$ and the same is used across all setting.}
    \label{fig:benchmark}
\end{figure}

\begin{figure}[ht] 
    \centering
    \includegraphics{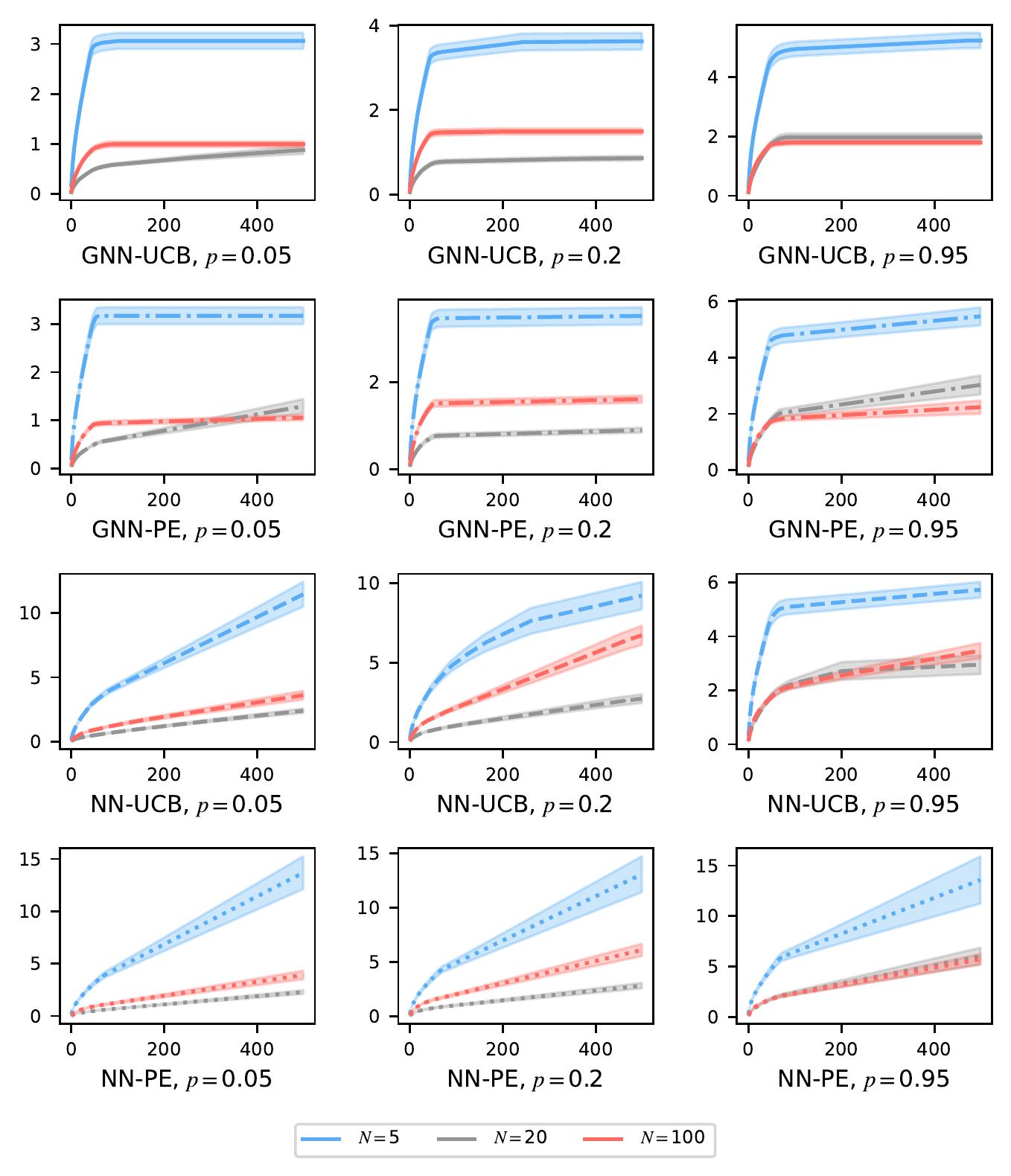}
    \caption{Effect of graph size on performance of \gnnus, \gnnucb, \nnus, and \nnucb. 
    GNN methods perform well regardless of value of $N$. 
    Inference on sparse small graphs is challenging since the random graphs tend to have very few edges.}
    \label{fig:scalability}
\end{figure}

\begin{figure}[ht] 
    \centering
    \includegraphics{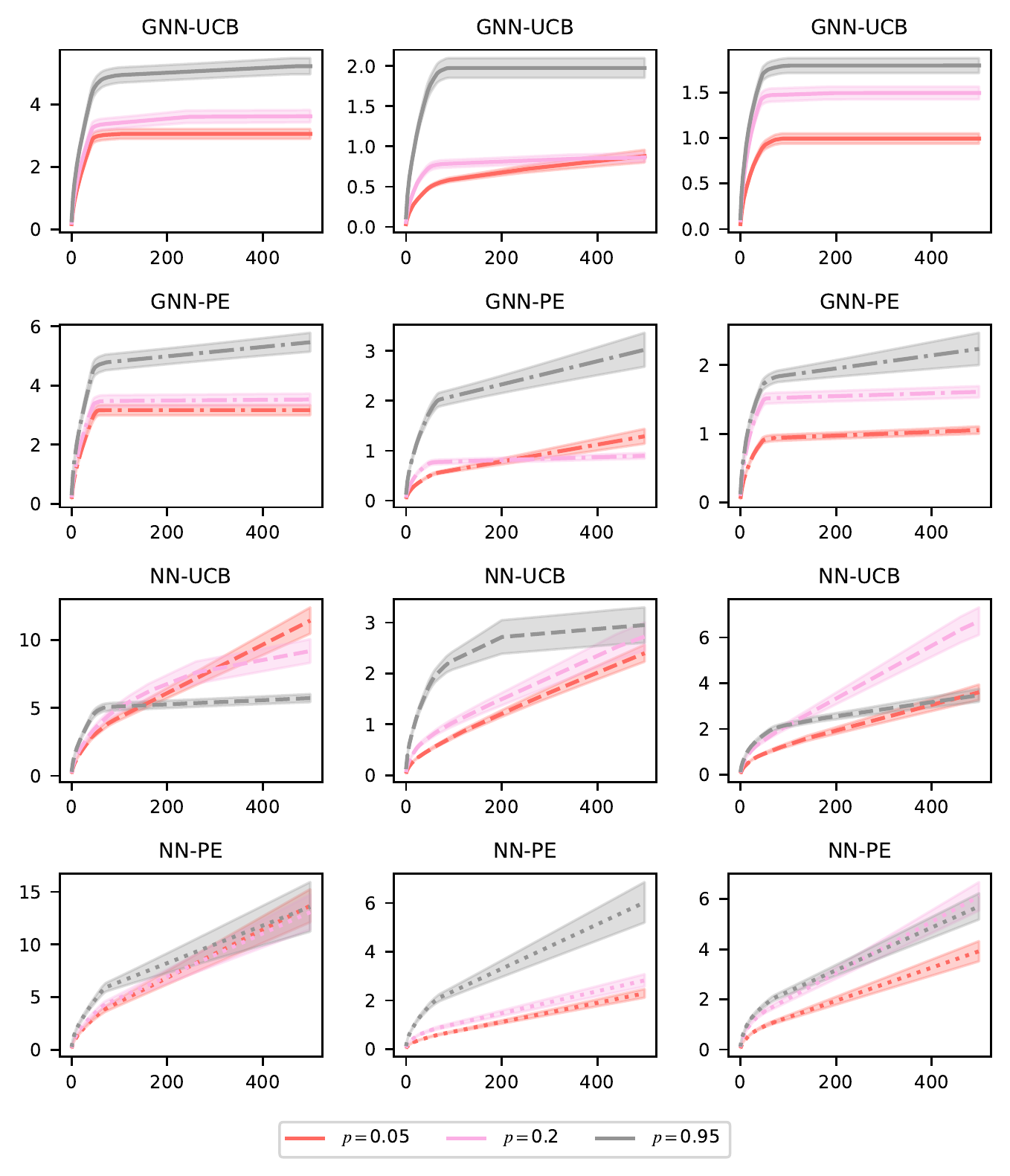}
    \caption{Effect of edge density on performance of \gnnus, \gnnucb, \nnus, and \nnucb. The NN algorithms tend to improve as $p$ grows.}
    \label{fig:density}
\end{figure}


\end{document}